%% file: FG_DDF_full_root.tex
\documentclass[journal]{IEEEtran}
\IEEEoverridecommandlockouts
\usepackage[utf8]{inputenc}
\usepackage{textcomp}
\usepackage{graphicx}
\usepackage[version=4]{mhchem}
\usepackage{siunitx}
\usepackage{longtable,tabularx}
\usepackage{accents}
\usepackage{comment}
\usepackage{mathtools}
\usepackage{subfigure}
\usepackage{caption}
\usepackage{color}
\usepackage{amsmath,amssymb}
\usepackage{booktabs}
\usepackage{tabu}
\usepackage{algorithm}
\usepackage{algpseudocode}
\usepackage[hidelinks]{hyperref}
\usepackage{soul}
\usepackage[dvipsnames]{xcolor}
\usepackage{multirow}
\usepackage{enumerate}
\usepackage{svg}
\usepackage{epstopdf}
\usepackage{cite}

\usepackage{tikz}
\usepackage{siunitx} 
\usetikzlibrary{arrows, arrows.meta}
\usetikzlibrary{graphs}
\usetikzlibrary{snakes,backgrounds,patterns,matrix,shapes,fit,calc,shadows,plotmarks} 
\usetikzlibrary{bayesnet}

\usepackage{standalone}

\tikzstyle{arrow}=[arrows={{Latex[scale=0.5]}-}, thick]  

\tikzset{
    between/.style args={#1 and #2}{
         at = ($(#1)!0.5!(#2)$)
    }
}

\newcommand{\etal}{\emph{et al. }}

\def\BibTeX{{\rm B\kern-.05em{\sc i\kern-.025em b}\kern-.08em
    T\kern-.1667em\lower.7ex\hbox{E}\kern-.125emX}}
\begin{document}

\title{Scalable Factor Graph-Based Heterogeneous Bayesian DDF for Dynamic Systems \\
}

\author{Ofer~Dagan,~\IEEEmembership{Student Member,~IEEE,} Tycho~L.~Cinquini, Nisar~R.~Ahmed,~\IEEEmembership{Member,~IEEE}
\thanks{Manuscript received Month day, 2023; revised Month day, 2024.}
\thanks{The authors are with the Smead Aerospace Engineering Sciences Department, University of Colorado Boulder, Boulder, CO 80309 USA (e-mail: ofer.dagan@colorado.edu; Nisar.Ahmed@colorado.edu). }}%



\maketitle

\begin{abstract}
Heterogeneous Bayesian decentralized data fusion captures the set of problems in which two robots must combine two probability density functions over non-equal, but overlapping sets of random variables.
In the context of multi-robot dynamic systems, this enables robots to take a `divide and conquer' approach to reason and share data over complementary tasks instead of over the full joint state space.
For example, in a target tracking application, this allows robots to track different subsets of targets and share data on only common targets. 
This paper presents a framework by which robots can each use a local factor graph to represent relevant partitions of a complex global joint probability distribution, thus allowing them to avoid reasoning over the entirety of a more complex model and saving communication as well as computation costs.
From a theoretical point of view, this paper makes contributions by casting the heterogeneous decentralized fusion problem in terms of a factor graph, analyzing the challenges that arise due to dynamic filtering, and then developing a new conservative filtering algorithm that ensures statistical correctness.
From a practical point of view, we show how this framework can be used to represent different multi-robot applications and then test it with simulations and hardware experiments to validate and demonstrate its statistical conservativeness, applicability, and robustness to real-world challenges.

\end{abstract}




\section{Introduction}
\input{Text/0_Introduction.tex}

\section{The heterogeneous DDF problem}
\label{sec:hetroFusion}
\input{Text/2_hetroFusion}

\subsection{Related Work}
\label{subsec:relatedWork}

\input{Text/2p1_Related_work}

\section{Heterogeneous DDF using Factor Graphs}
\label{sec:FG_DDF} 
\input{Text/Background.tex}

\input{Text/5_FG_DDF}



\section{Experiments}
\label{sec:experiments}
\input{Text/6_Experiments}

\section{Conclusions}
\label{sec:conclusions}
\input{Text/7_Conclusions}

\newpage

\appendices
\section{Probabilistic Operations on Factor Graphs}
\label{sec:appendix}
\input{Text/Appendix.tex}



\bibliographystyle{IEEEtran}
\bibliography{references.bib}

\begin{IEEEbiography}[{\includegraphics[width=1.1in,height=1.25in,clip,keepaspectratio]{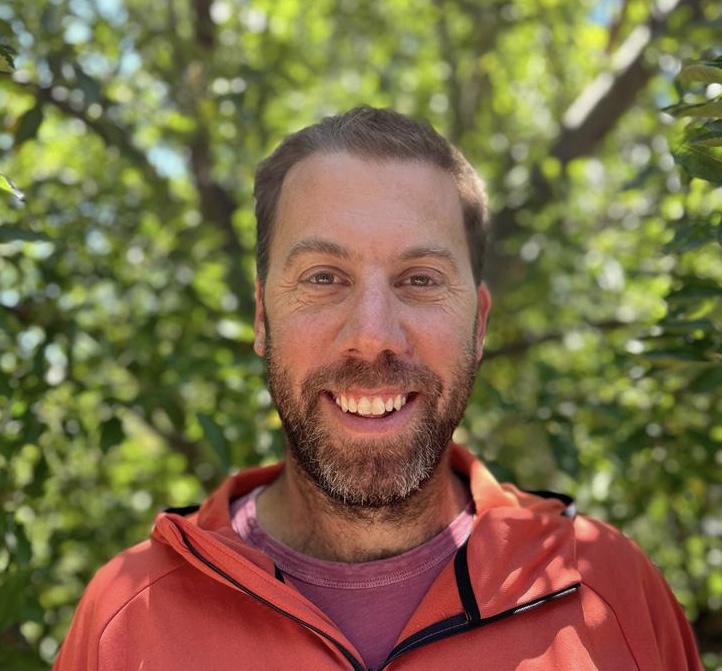}}]{Ofer Dagan}
received the B.S. degree in aerospace engineering, in 2010, and the M.S. degree in mechanical engineering, in 2015, from the Technion - Israel Institute of Technology, Haifa, Israel, and the Ph.d. degree
in aerospace engineering with the Ann and H.J.
Smead Aerospace Engineering Sciences Department,
University of Colorado Boulder, Boulder, CO, USA in 2024. 
He is currently a postdoctoral fellow at the Autonomous Decision and Control Lab (ADCL) at the University of Colorado Boulder.
From 2010 to 2018 he was a research engineer in the aerospace industry. 
His research interests include theory and algorithms for decentralized Bayesian reasoning in heterogeneous autonomous systems.
\end{IEEEbiography}

\begin{IEEEbiography}[{\includegraphics[width=1.1in,height=1.25in,clip,keepaspectratio]{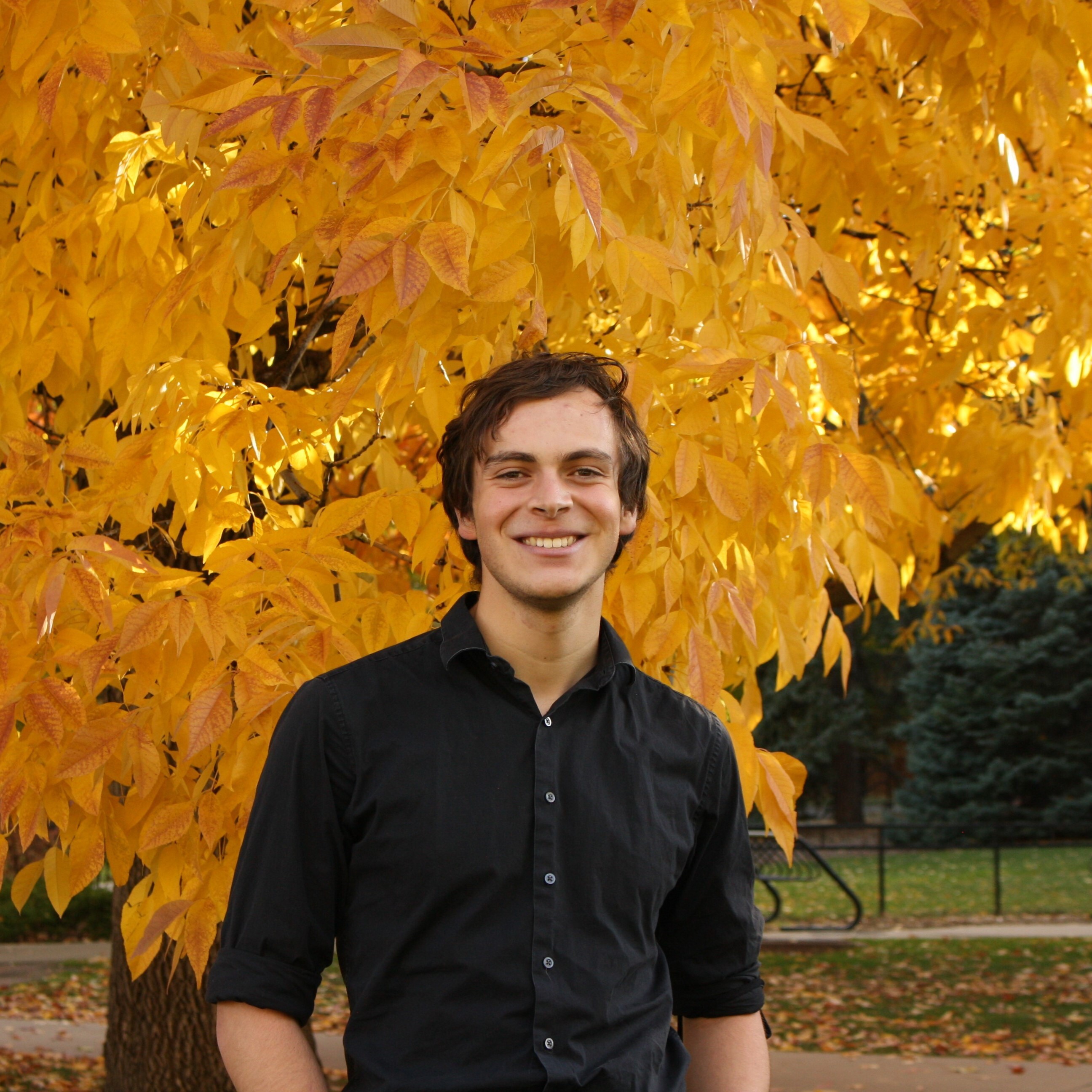}}]{Tycho L. Cinquini}
 received the B.S. degree in aerospace engineering in 2023 from the Ann and H.J. Smead Aerospace Engineering Sciences Department, University of Colorado Boulder, Boulder, CO, USA. He is currently working toward the M.S. degree in aerospace engineering with a focus on autonomous systems through the same college. His research interests include the implementation of various algorithms on multi-robot hardware systems.
\end{IEEEbiography}

\begin{IEEEbiography}[{\includegraphics[width=1.1in,height=1.25in,clip,keepaspectratio]{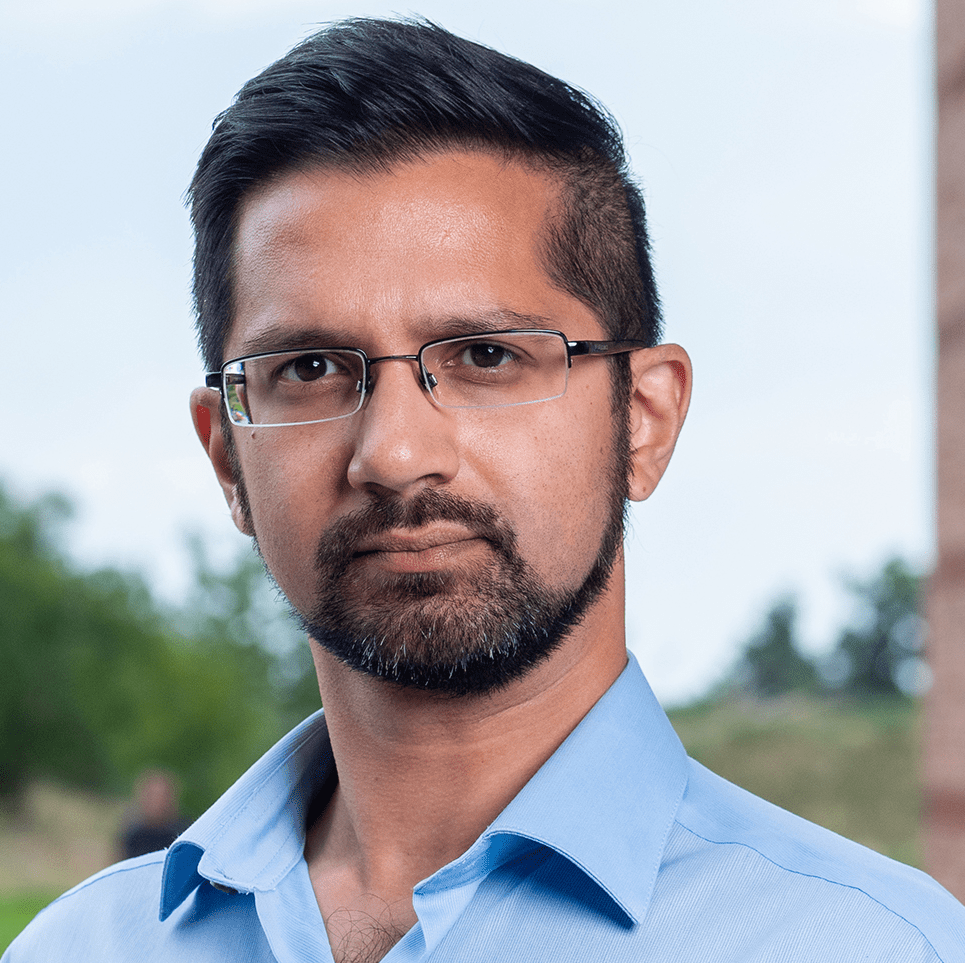}}]{Nisar R. Ahmed}
received the B.S. degree in engineering from Cooper Union, New York City, NY, USA, in 2006 and the Ph.D. degree in mechanical engineering from Cornell University, Ithaca, NY, USA,
in 2012.
He is an Associate Professor of Autonomous Systems and H. Joseph Smead Faculty Fellow with Ann and H.J. Smead Aerospace Engineering Sciences Department, University of Colorado Boulder, Boulder, CO, USA. He was also a Postdoctoral Research Associate with Cornell University
until 2014. His research interests include the development of probabilistic models and algorithms for cooperative intelligence in mixed human–machine teams.
\end{IEEEbiography}

\end{document}

%% file: Text/0_Introduction.tex
The idea of a team of autonomous agents (robots) cooperating on a joint task can be allegorized to a group of people working together. Often people have different capabilities, different knowledge, and different worldviews. 
However, when collaborating, they naturally know how to summarize only the relevant information to achieve a joint goal.
For a team of robots that needs to work together, this human capability is not trivial. 
The robot's ability to make sense and act in a constantly changing environment is much less effective in this multi-perspective aspect than what the human brain does.
One of the main approaches to allow robots to reason about their uncertainty is the probabilistic approach \cite{thrun_probabilistic_2005}, where a robot models the uncertainty in how it perceives the world using a probability density function (pdf). 

In Bayesian decentralized data fusion (DDF) this approach is leveraged to allow any two robots in a network of $n_r$ robots to gain new data by sharing their posterior pdfs, representing their local estimates. 
However, DDF methods do not scale well as the number of robots in the network increases since they frequently require all robots to process and communicate the full global pdf. 
Our work enables robots to exploit the probabilistic conditional independence structure, inherent in many robotic applications to ‘break’ the global problem into smaller, locally relevant problems, thus significantly improving communication and computation requirements for each robot. 
For this reason, this work leverages factor graphs, one of the most general frameworks for analyzing probabilistic conditional independence structure and efficient inference \cite{frey_factor_1997, dellaert_factor_2021}, to explore the sub-class of DDF problems, defined as heterogeneous DDF \cite{dagan_exact_2023}.
We develop a novel framework for heterogeneous DDF problems dubbed FG-DDF, and
show how FG-DDF allows for representation, rigorous analysis, and the solution of heterogeneous DDF problems directly on the graph. We demonstrate its performance, applicability, and robustness with simulations and hardware experiments on robotic platforms.
More specifically, our contributions are:
\begin{enumerate}
    \item Development of the FG-DDF framework, which leverages factor graphs to enable robots to break down a global problem into smaller locally relevant problems. 
    By using factor graphs to represent the relevant partitions of the joint probability distribution, robots can significantly reduce communication and computation requirements. 
    \item Analysis of the challenges that arise due to filtering in heterogeneous fusion systems and development of the conservative filtering algorithm. 
    The algorithm addresses the challenges of maintaining the conditional independence structure and avoiding double counting of common data in heterogeneous fusion. 
    \item Exploration of methods for accounting for common data in Bayesian decentralized data fusion and formulation into a heterogeneous extension to the (homogeneous) covariance intersection (CI) algorithm, dubbed HS-CI, to allow for cyclic network topologies. 
    \item Validation of the effectiveness of the FG-DDF framework through simulations and hardware experiments on robotic platforms in real-world scenarios.
    These experiments demonstrate the applicability, robustness, and scalability of the approach and show that under various conditions the system provides consistent and conservative estimates.
    
\end{enumerate}   

\textbf{Novelty with respect to previous work} \cite{dagan_factor_2021,dagan_conservative_2022,dagan_non-linear_2023}:
This paper evaluates and brings to maturity our work on the FG-DDF framework for heterogeneous Bayesian DDF. 
To develop the theory behind FG-DDF, our previous work (\cite{dagan_factor_2021,dagan_conservative_2022}) made some limiting assumptions, such as acyclic network topologies, perfect communication, and linear transition and observation models. 
This paper aims to explore 
\textit{how robust and applicable is the FG-DDF to real-world scenarios?} 
It builds on our previous work and expands it by relaxing the above assumptions and testing the performance of the FG-DDF framework with large-scale cyclic network topologies, $50\%$ message dropouts, non-linear transition and observation models, and measurement outliers. 
These are tested in simulations and hardware experiments of multi-robot multi-target tracking and cooperative localization applications.
We show that FG-DDF is robust, scalable, and applicable to real-world scenarios.

The rest of the paper is structured as follows: Sec. \ref{sec:hetroFusion} presents the heterogeneous DDF problem, the key technical challenges pertaining to the problem, and discusses relevant existing work. 
Sec. \ref{sec:FG_DDF} includes the main theoretical contributions for developing a factor graph-based framework for heterogeneous DDF. 
Sec. \ref{sec:algorithm} then puts it all together to present the FG-DDF algorithm and our conservative filtering approach.
In Sec. \ref{sec:experiments} we test the proposed framework and algorithms in simulation and hardware experiments and analyze its performance under realistic conditions. Sec. \ref{sec:conclusions} then draws conclusions and discusses future directions.


%% file: Text/2_hetroFusion.tex
Let $V$ be a global set of random variables (rvs) describing states of interest monitored by a set $N_r$ of $n_r$ autonomous robots. 
The states of interest are distributed between the robots such that each robot $i$ monitors some ``local states of interest", which are a subset of the global set $\chi^i\subseteq V$. 
The set $\chi^i$ can be divided into a set of local states $\chi^i_L$, which are not monitored (observed) by any other robot in the network, and a set of common states between robot $i$ and its neighbors $\chi^i_C=\bigcup_{j\in N_r^i}^{}\chi^{ij}_C$, where $N^i_{r}\subseteq N_r$ is the set of robots communicating with robot $i$ and $\chi^i=\chi^i_L\bigcup \chi^i_C$.

In Bayesian DDF, each robot $i$ is an independent entity, collecting data over a set of random variables of interest $\chi^i$ from its local sensors and via communication with neighboring robots. 
A robot can update its local prior pdf over $\chi^i$ by Bayesian fusion of:
(i) independent local measurements $y^{i,l}_{k}\in Y^i_k$, described by the conditional likelihood $p(Y^i_k|\chi^i_k)=\prod_{l}p(y^{i,l}_{k}|\chi^{i,l}_{k})$, where $\chi^{i,l}_{k}$ is the subset of states, measured by the $l$ measurement $y^{i,l}_{k}$, taken by robot $i$ at time step $k$;
(ii) a posterior pdf received from any neighboring robot $j\in N^i_r$ via the peer-to-peer heterogeneous fusion rule \cite{dagan_exact_2023}, 
\begin{equation}
    \begin{split}
        p_f^i(\chi^i|&Z^{i,+}_k)\propto \\
        &\frac{p^i(\chi^{ij}_C|Z^{i,-}_k)p^j(\chi^{ij}_C|Z^{j,-}_k)}{p^{ij}_c(\chi^{ij}_C|Z^{i,-}_k \cap Z^{j,-}_k)}  
         \cdot p^i(\chi^{i\backslash j}|\chi^{ij}_C,Z^{i,-}_k).
    \end{split}
    \label{eq:Heterogeneous_fusion}
\end{equation}
Here $\chi^{i\setminus j}$ is the set of non-mutual rvs to $i$ and $j$, where `$\setminus$' is the set exclusion operation, $Z^{i,-}_k$ and $Z^{i,+}_k$ are the data sets available for robot $i$ at time $k$ prior and post fusion, respectively. 
$p_f^i(\cdot)$ is the fused posterior pdf at robot $i$, $p^{ij}_c(\chi^{ij}_C|Z^{i,-}_k \cap Z^{j,-}_k)$ is the pdf over robots $i$ and $j$ common rvs, given their common data.
Note that when $\chi^i=\chi^j=\chi_C^{ij}$ and $\chi^{i\backslash j}=\varnothing$, the equation degenerates to the classic homogeneous Bayesian fusion rule \cite{chong_distributed_1983}. 
But, in the more general case, where two robots $i$ and $j$ hold distributions over overlapping sets of variables, i.e., $\chi^i\cap \chi^j\neq\varnothing$ and $\chi^i\setminus \chi^j\neq\varnothing$, (\ref{eq:Heterogeneous_fusion}) defines the \emph{Heterogeneous State} (HS) fusion equation \cite{dagan_exact_2023}. 
However, (\ref{eq:Heterogeneous_fusion}) is only valid if the sets of non-mutual variables held by each robot are conditionally independent given the common set, that is $\chi^{i\setminus j}\perp \chi^{j\setminus i}|\chi^{ij}_C$ \cite{dagan_exact_2023}. 
As it turns out, in many cooperative robotics problems, this conditional independence structure inherently exists and can be exploited for heterogeneous fusion.  
However, there are two main challenges pertaining to the heterogeneous DDF problem:
\begin{enumerate}
    \item \textbf{Maintaining the conditional independence structure:}
     In dynamic problems, recursive solutions such as filtering, require marginalization of past variables. 
    This marginalization during filtering breaks the conditional independence between the non-mutual variables.
    For example, the graph in Fig. \ref{fig:fullGraph}(a) encodes the local pdf factorization held by robot $i$ in relation to two other robots j and m,
    \begin{equation}
        \begin{split}
            &p(\chi^i_{2:1}|Z^{i,+}_1) = p(\chi_L^i)\cdot \\   &p(\chi^{ijm}_{C,1}|\chi_L^i)\cdot p(\chi^{im\setminus j}_{C,1}|\chi^{ijm}_{C,1},\chi_L^i) \cdot p(\chi^{ij\setminus m}_{C,1}|\chi^{ijm}_{C,1},\chi_L^i)\cdot \\
             &p(\chi^{ijm}_{C,2}|\chi^{ijm}_{C,1}) 
            \cdot p(\chi^{im\setminus j}_{C,2}|\chi^{im\setminus j}_{C,1})   \cdot p(\chi^{ij\setminus m}_{C,2}|\chi^{ij\setminus m}_{C,1}), 
        \end{split}
        \label{eq:factorization}
    \end{equation}
    where the common variables are separated into three different sets $\chi_C^{ijm}$, $\chi_C^{ij\setminus m}$ and $\chi_C^{im\setminus j}$, representing variables common to the three robots, variables common to $i$ and $j$ but not to $m$ and similarly variables common to $i$ and $m$ but not to $j$, respectively. 
    Here the conditioning on the data $Z_1^{i,+}$ is omitted from the right side of the equations for brevity. 
    Marginalizing the rvs of time step 1 results in a coupled, dense graph (Fig. \ref{fig:fullGraph}(b)), corresponding to, 
    \begin{equation}
        \begin{split}
            p(\chi^i_{2}|Z^{i,+}_1)= \int p(\chi^i_{2:1}|Z^{i,+}_1)d\chi_{C,1}^i,
        \end{split}
        \label{eq:approx_definition}
    \end{equation}
    
    \item \textbf{Avoiding double counting of common data:}
    The common data between robots, is expressed by $p^{ij}_c(\chi^{ij}_C|Z^{i,-}_k \cap Z^{j,-}_k)$, in the denominator of (\ref{eq:Heterogeneous_fusion}). 
    That is common a priori knowledge that was already fused into robots’ model of the world (pdf) and exchanged knowledge that has circulated back through the network to both robots. 
    This common data must be computed and removed, to guarantee that new data are treated as such only once. 
\end{enumerate}

\input{Figures/Full_graph_fig.tex}

The heterogeneous DDF problem considered in this paper is therefore enabling the analysis and solution of dynamic heterogeneous multi-robot systems, such that it results in a conservative posterior pdf while maintaining the conditional independence requirement to allow for heterogeneous fusion.

%% file: Figures/Full_graph_fig.tex
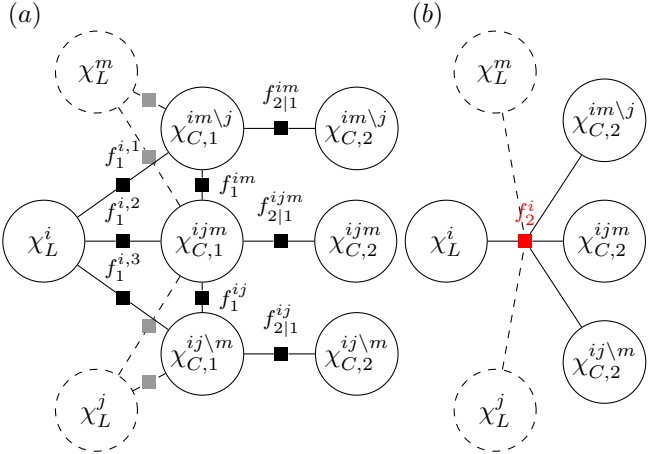
\begin{figure}[tb]
\scalebox{0.93}{}{%
\begin{tikzpicture}[ new set=import nodes]
 \begin{scope}[nodes={set=import nodes}]
      
      \node (a) at (-0.35,3) {$(a)$};
      \node (x_Li)[latent, minimum size=31pt] at (-0.1,0) {$\chi^i_L$};
      \node (x_Lm)[latent,dashed, minimum size=31pt] at (0.6,2.25) {$\chi^m_L$};
      \node (x_Lj)[latent,dashed, minimum size=31pt] at (0.6,-2.25) {$\chi^j_L$};
      \node [latent, right=of x_Li, yshift=+1.5cm, minimum size=31pt] (x_im) {$\chi^{im\setminus j}_{C,1}$};
      \node [latent, right=of x_Li, yshift=0, minimum size=31pt] (x_ijm) {$\chi^{ijm}_{C,1}$};
      \node [latent, right=of x_Li, yshift=-1.5cm, minimum size=31pt] (x_ij) {$\chi^{ij\setminus m}_{C,1}$};
      \node [factor, between=x_Li and x_im,label=$f^{i,1}_{1}$ ] (yi11) {};
      \node [factor, between=x_Li and x_ijm,label=$f^{i,2}_{1}$ ] (yi21) {};
      \node [factor, between=x_Li and x_ij,label=$f^{i,3}_{1}$ ] (yi31) {};
     
      \node [factor, between=x_im and x_ijm ,label=right:$f^{im}_1$] (f_im) {};
      \node [factor, between=x_ij and x_ijm ,label=right:$f^{ij}_1$] (f_ij) {};
     
     \node [factor, right=of x_im,  label={$f^{im}_{2|1}$}] (f2_im) {};
      \node [latent, right=of f2_im, xshift=-0.65cm, minimum size=31pt] (x2_im) {$\chi^{im\setminus j}_{C,2}$};
      
      \node [factor, right=of x_ijm,   label={$f^{ijm}_{2|1}$}] (f2_ijm) {};
      \node [latent, right=of f2_ijm, xshift=-0.65cm, minimum size=31pt] (x2_ijm) {$\chi^{ijm}_{C,2}$};
      
      \node [factor, right=of x_ij, label={$f^{ij}_{2|1}$}] (f2_ij) {};
      \node [latent, right=of f2_ij, xshift=-0.65cm, minimum size=31pt] (x2_ij) {$\chi^{ij\setminus m}_{C,2}$};
      
      \node [factor,fill=black!40, between=x_Lj and x_ijm] (f_j11) {};
      \node [factor,fill=black!40, between=x_Lj and x_ij] (f_j21) {};
      
      \node [factor,fill=black!40, between=x_Lm and x_ijm] (f_m11) {};
      \node [factor,fill=black!40, between=x_Lm and x_im] (f_m21) {};
      
      
      \node (b) at (5,3) {$(b)$};
      \node (x_Lib)[latent, minimum size=31pt] at (5.25,0) {$\chi^i_L$};
      \node (x_Lmb)[latent,dashed, minimum size=31pt] at (5.9,2.25) {$\chi^m_L$};
      \node (x_Ljb)[latent,dashed, minimum size=31pt] at (5.9,-2.25) {$\chi^j_L$};

      \node [latent, right=of x_Lib, xshift=0.0cm, minimum size=31pt] (x2_ijmb) {$\chi^{ijm}_{C,2}$};
      
      \node [latent, above=of x2_ijmb, yshift=-0.5cm, minimum size=31pt] (x2_imb) {$\chi^{im\setminus j}_{C,2}$};
      
      \node [latent, below=of x2_ijmb, yshift=0.5cm, minimum size=31pt] (x2_ijb) {$\chi^{ij\setminus m}_{C,2}$};
     
     \node [factor,fill=red!100, between=x_Lib and x2_ijmb, label=\textcolor{red!100}{{$f^{i}_2$}} ] (f2b) {};
      
  \end{scope}
  
 \graph {
    (import nodes);
   
    
   
    {x_im,x_Li}--yi11, 
    {x_ijm,x_Li}--yi21,  
    {x_ij,x_Li}--yi31,
    {x_ij,x_ijm}--f_ij,
    {x_im,x_ijm}--f_im,
    {x2_ij,x_ij}--f2_ij,
    {x2_im,x_im}--f2_im,
    {x2_ijm,x_ijm}--f2_ijm,
    {x_Lm}--[dashed]f_m21, 
    {x_Lm}--[dashed]f_m11,
    {x_Lj}--[dashed]f_j21, 
    {x_Lj}--[dashed]f_j11,
    {x_im}--[dashed]f_m21,
    {x_ijm}--[dashed]f_m11,
    {x_ij}--[dashed]f_j21, 
    {x_ijm}--[dashed]f_j11,
    {x_Lib,x2_imb, x2_ijb, x2_ijmb }--f2b,
    {x_Ljb,x_Lmb}--[dashed]f2b,
   
    };
    
\end{tikzpicture}}
\caption{Neighborhood graph perspective: factor graph representing robot $i$' local pdf with hidden local variables of neighboring robots $m$ and $j$. Dashed nodes and grey factors are hidden from robot $i$. (a) graph before marginalization of time step 1, demonstrating conditional independence structure (b) fully connected graph after marginalization. }
      \label{fig:fullGraph}
\end{figure}

%% file: Text/2p1_Related_work.tex
The conditional independence structure and the flow of data in the system are both affected by marginalization, as it introduces dependencies between previously independent variables by `marrying parents' (moralization) in the corresponding probabilistic graphical model \cite{cowell_specification_2007}. 
In state estimation heterogeneous--fusion problems, marginalization occurs (i) in filtering, and (ii) when a robot sends a message over the marginal of common variables. 
We can then categorize heterogeneous fusion algorithms based on: the solution approach, that is, if the inference is done over the full-time history (as in smoothing), or recursive (as in filtering); and the type of the fused rvs (static/dynamic), respectively.  
Table \ref{tab:related_work} categorizes selected existing heterogeneous fusion algorithms with respect to these two dimensions of the underlying problem.

\begin{table*}[tb]
\renewcommand{\arraystretch}{1.5}
\caption{Categorization of PGM-based fusion methods.  }
    \centering
    \begin{tabular}{c|c|c|c}
%
        \multirow{2}{*}{\textbf{Fused rvs / Inference solution}}  & \multirow{2}{*}{\textbf{Static}} & \multicolumn{2}{c}{\textbf{Dynamic}} \\
        & & \textbf{Smoothing} & \textbf{Recursive (filtering)}   \\ \hline
        All - $\chi$ (homogeneous) & Makarenko \etal \cite{makarenko_decentralised_2009}  & Makarenko \etal \cite{makarenko_decentralised_2009}  & ---  \\ \hline
        All dynamic - $\chi_D$  & N/A  & ---  & ---  \\ \hline
        All static- $\chi_S$  & ---  & ---  & ---   \\ \hline
        Subset of dynamic - $\chi_d$  &  N/A & Etzlinger \etal \cite{etzlinger_cooperative_2017}  & ---   \\ \hline
        Subset of static - $\chi_s$  & Paskin \etal \cite{paskin_robust_2004}  & Cunningham \etal \cite{cunningham_ddf-sam_2013}  & ---   \\ \hline
    \end{tabular}
    \label{tab:related_work}
\end{table*} 

\textbf{Static problems:} 
In \cite{paskin_robust_2004} Paskin \etal presents a distributed inference problem in a network of static sensors. 
It is solved by using a robust message-passing algorithm on a junction tree (JT), but this is limited to static variables and requires the full construction of the tree before performing inference. 
In \cite{makarenko_decentralised_2009} Makarenko \etal formulates Paskin's JT algorithm as a DDF problem and extends it such that robots can fuse either static or dynamic states. 
They show that for static network topology and a static model, the decentralized JT (D-JT) algorithm is equivalent to the channel filter (CF) \cite{grime_data_1994} and the Hugin \cite{andersen_hugin_1989} algorithms. 
However, the algorithm is limited to a single common state, i.e., the focus is on homogeneous problems and it can not be used to solve heterogeneous problems. 

\textbf{Dynamic problems:}
While the D-JT algorithm by Makarenko \etal can solve dynamic problems, it does not suggest a solution for recursive (filtering) systems and, thus can be only used when the states are augmented in time, as in smoothing. 
Other decentralized algorithms that solve dynamic smoothing problems are DDF-SAM $2.0$ by Cunningham \etal \cite{cunningham_ddf-sam_2013} and CoSLAS by Etzlinger \etal \cite{etzlinger_cooperative_2017}. 
In DDF-SAM $2.0$, the authors use a factor graph to solve a multi-robot SLAM problem.
Robots estimate their full trajectory (poses) over time and share a (static) subset of their map, represented by landmarks. 
Similarly, the work in \cite{etzlinger_cooperative_2017} is based on a factor graph representation to solve a cooperative simultaneous localization and synchronization (CoSLAS) problem. 
In \cite{etzlinger_cooperative_2017}, robots share a subset of dynamic states, namely clock and position states, using a message-passing algorithm on the factor graph. 
This thesis generalizes their work in several aspects:
\begin{enumerate}
    \item A more flexible definition of neighbors -- in \cite{etzlinger_cooperative_2017} robots are considered neighbors only if they share a relative measurement. In this paper, robots can be defined as neighbors if they share common inference tasks.
    \item Application -- the algorithm in \cite{etzlinger_cooperative_2017} is tailored to the specific CoSLAS application, whereas here the problem definition allows for the representation and solution of different robotic problems (see Sec.\ref{sec:hetroFusion}).
    \item Robustness -- in \cite{etzlinger_cooperative_2017} the robots are dependent on each other to form an estimate, i.e., a robot cannot infer its state based on only locally available data. The DDF framework used in this paper allows for more robust and independent collaboration.
    \item Problem space -- as discussed and shown in Table \ref{tab:related_work}, \cite{etzlinger_cooperative_2017} do not suggest a time recursive solution, for a filtering scenario, where Sec. \ref{sec:algorithm} in this paper addresses this challenge.
\end{enumerate}


Another work worth mentioning here is the work by Chong and Mori \cite{chong_graphical_2004}, where they use both Bayesian networks and information graphs to identify conditional independence and track common information, respectively. 
While they use conditional independence to reduce state dimension, it is only for communication purposes and not for local state reduction. 
Since in this case the graphical model representation is used for analysis purposes and not for inference on the graph itself, it does not fit the categorization in Table \ref{tab:related_work}.

As can be seen in Table \ref{tab:related_work} and discussed above, there is a gap in the literature concerning methods that solve heterogeneous fusion problems in dynamical systems, more specifically for recursive inference.
This paper presents FG-DDF, a factor graph-based framework to analyze and solve recursive and non-recursive (i.e. storing all measurements) heterogeneous DDF problems. 
In FG-DDF, the variables to be shared can be dynamic, static, or both, which fills up the existing gap in theory and literature. 
To develop the theory, analyze conditional independence, and track the data flow in the network, we focus on linear models with Gaussian noise in undirectional acyclic network topologies.
However, the fundamental probabilistic operations derived here for FG-DDF could be extended to any other variety of dynamic probabilistic models. (Understand that you don't want to overclaim or overstate -- but it is fair to say that the theory is general as far as the PGM goes, even if the specific factor implementation is constrained to linear/Gaussian here...) 
We then relax these assumptions and demonstrate the applicability of the framework in realistic scenarios in large-scale problems, different network topologies, nonlinear models, message dropouts, and measurement outliers.
The theory and algorithms developed in this paper can then be used to allow scalable collaboration between robots that use different algorithms and enable heterogeneous teams. 
For example, in \cite{dagan_towards_2023}, we show how to use the FG-DDF framework for a heterogeneous SLAM and tracking system, where robots can run different (e.g., lidar visual inertial) SLAM algorithms. 

Note, that while in this paper, for implementation reasons we assume that factors are defined by the first two moments of the pdf, 
the fundamental probabilistic operations derived here for FG-DDF could be extended to any other variety of dynamic probabilistic models.

%% file: Text/Background.tex
\subsection{Factor Graphs and Conditional Independence}
In recent years factor graphs \cite{frey_factor_1997} have been used to study and solve a variety of robotic applications \cite{dellaert_factor_2021}. Factor graphs are arguably the most general framework to analyze and express conditional independence, as such, they directly express the sparse structure of decentralized problems. 
 
A factor graph is an undirected bipartite graph $\mathcal{F}=(U,V,E)$ that represents a function, proportional to the joint pdf over all random variable nodes $v_m\in V$, and factorized into smaller functions given by the factor nodes $f_l\in U$. An edge $e_{lm}\in E$ in the graph can only connect a factor node \emph{l} to a variable node \emph{m}.
The joint distribution over the graph is then proportional to the global function $f(V)$:
\begin{equation}
    p(V)\propto f(V)=\prod_{l}f_l(V_l),
    \label{eq:factorization}
\end{equation}
where $f_l(V_l)$ is a function of only those variables $v_m\in V_l$ connected to the factor \emph{l}, thus making the factorization of the joint pdf easy to directly read from the graph.

The product representation of the factor graph $\mathcal{F}$ can be transformed into summation by converting the factors to log space (see \cite{koller_probabilistic_2009}).
A special case of interest is the canonical (information) form of the Gaussian distribution, as it is tightly connected to the factor graph, representing the underlying distribution. 
In this case, both the graph and the joint information matrix represent the conditional independence structure of the distribution - the graph from its factorization, and the matrix with zero off-diagonal terms between conditionally independent variables. 
In the canonical (information) form of the Gaussian distribution, factors are expressed by two elements, the information vector ($\zeta$), and the information matrix ($\Lambda$).
The direct sum ($\oplus$) of all factors in the graph describes the multivariate Gaussian distribution in canonical form,
\begin{equation}
    p(V)\sim \mathcal{N}^{-1}(V;\zeta, \Lambda)= \bigoplus_{l}f_l(V_l),
    \label{eq:canonicalFactorization}
\end{equation}
with $\mathcal{N}^{-1}$ denoting the information parameterization of the Gaussian distribution $\mathcal{N}$ \cite{schon_manipulating_2011}.
Here, all factors are Gaussian distributions represented in the canonical form $f_l(V_l)\sim \mathcal{N}^{-1}(V_l;\zeta_l, \Lambda_l)$ of the Gaussian distribution $\mathcal{N}(V_l;\mu_l, \Sigma_l)$, having a mean $\mu_l=\Lambda_l^{-1}\zeta_l$ and covariance $\Sigma_l=\Lambda_l^{-1}$.

\begin{table}[tb]
\renewcommand{\arraystretch}{1.75}
\caption{Factors dictionary, giving examples for different types of factors, their notation, and their pdf interpretation.  }
    \begin{center}
    \begin{tabular}{c|c|c}
        Factor  & Type & Proportional to  \\ \hline
        $f^{i,l}_k$ & Local measurement & $p(y^{i,l}_{k}|\chi^{i,l}_{k})$ \\ \hline
        $f^{ij\setminus m}_{2|1}$ & Dynamic prediction   &  $p(\chi^{ij\setminus m}_{C,2}|\chi^{ij\setminus m}_{C,1})$ \\ \hline
        $f_k^{ij}$ & Fusion & $p(\chi^{ij}_C|Z^{j,-}_k)$ (from (\ref{eq:Heterogeneous_fusion}))\\ \hline
         \textcolor{red}{$f^{i}_2$}& Dense marginalization & $\int p(\chi^i_{2:1}|Z^{i,+}_1)d\chi_{C,1}^i$  \\ \hline
         \textcolor{red}{$\tilde{f}^{ij}_2$}& Approximate marginalization & $\int \tilde{p}(\chi^i_{2:1}|Z^{i,+}_1)d\chi_{C,1}^i$ (\ref{eq:local_k21})\\ 
    \end{tabular}
    \end{center}
    \label{tab:factor_definitions}
    \vspace{-0.2in}
\end{table} 

%% file: Text/5_FG_DDF.tex
\begin{figure}[bt] 
    \centering
    \includegraphics[width=0.48\textwidth]{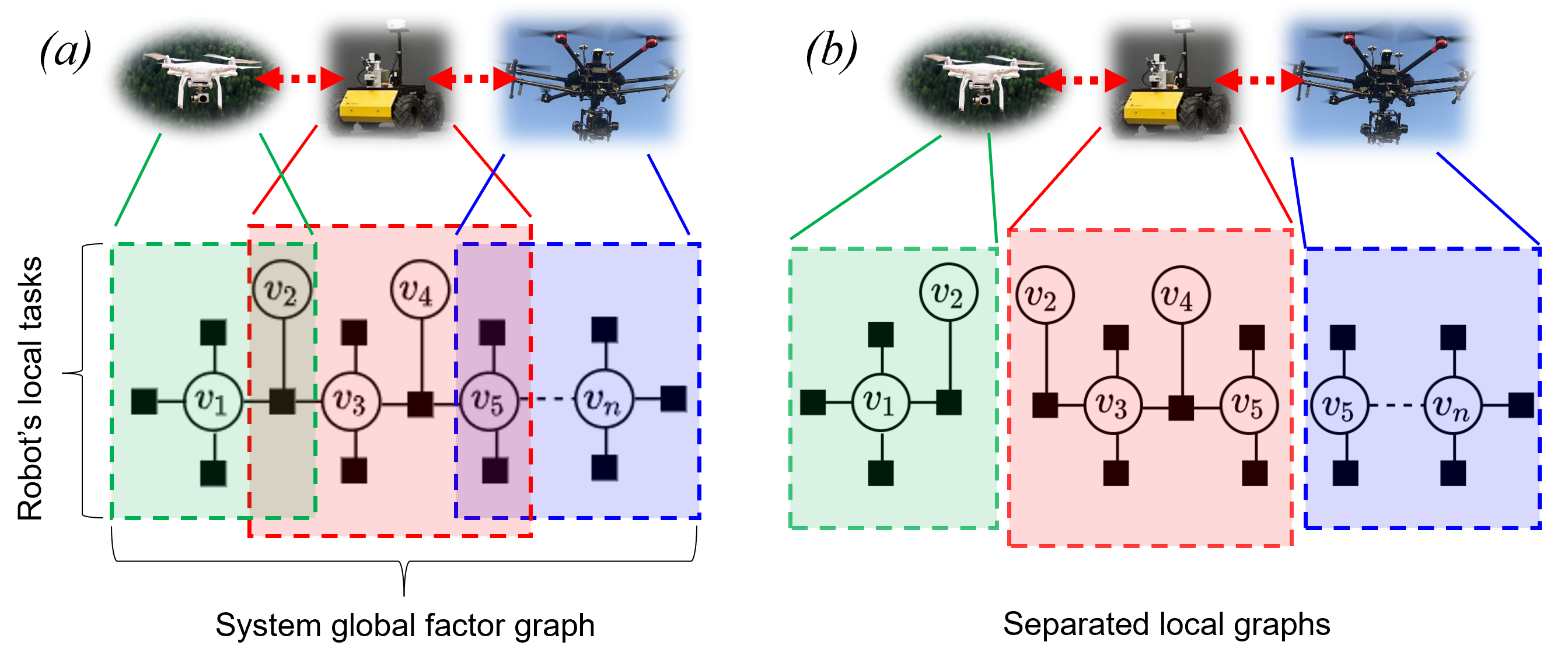}
    \caption{Shift in approach from requiring all robots to reason over the full global graph, irrespective of their smaller local tasks (a), to each robot reasons over its local smaller graph, representing its task (b).   }
    \label{fig:FG_approach}
    \vspace{-0.2in}
\end{figure}

The main shift in approach to factor graph-based decentralized data fusion (FG-DDF) is the use of conditional independence to split the full, \emph{global}, system graph into smaller \emph{local} sub-graphs. 
Each robot maintains and reasons over a smaller problem, representing its local inference task, instead of maintaining the full graph, representing the global inference task (Fig. \ref{fig:FG_approach}).  

In the next sections, we show how to exploit the factor graph's ability to naturally represent the conditional independence structure to solve the heterogeneous DDF challenges described in Sec. \ref{sec:hetroFusion}. 
We start by formulating the DDF problem as a factor graph, where fusion can be viewed as exchanging factors between robots, and describe how to avoid the double counting problem. 
Then we show how to manipulate the graph in order to maintain conditional independence and guarantee conservative fusion.


\subsection{DDF with Factor Graphs} 
Consider two robots $i$ and $j$, tasked with inferring the values of two overlapping subsets of the global set of rvs $\chi$, such that,
\begin{equation}
    \chi=\chi^i\cup \chi^j=\chi^{i\backslash j}\cup \chi^{j\backslash i} \cup \chi^{ij}_C,
    \label{eq:sets_Definitions}
\end{equation}
where, as before, $\chi^{i\backslash j}$ and $\chi^{j\backslash i}$ are mutually exclusive. 
From the global factor graph (Fig. \ref{fig:fusion}(a)), representing the joint pdf over $\chi$, it can be seen that these two subsets are conditionally independent given the common subset, i.e. $\chi^{i\backslash j} \perp \chi^{j\backslash i}|\chi^{ij}_C$. 
We leverage this conditional independence to split the graph into two subgraphs, held by each robot, representing the pdf over their local subset of variables, as seen in Fig. \ref{fig:fusion}(a)-(b).


\input{Figures/newFgDDF}

We can rearrange the terms in the heterogeneous fusion rule given in (\ref{eq:Heterogeneous_fusion}), 
\begin{equation}
    \begin{split}
        p_f^i(\chi^i|Z^{i,+}_k)\propto p^i(\chi^{i}|Z^{i,-}_k) \cdot \frac{p^j(\chi^{ij}_C|Z^{j,-}_k)}{p^{ij}_c(\chi^{ij}_C|Z^{i,-}_k \cap Z^{j,-}_k)},  
    \end{split}
    \label{eq:Heterogeneous_fusion2}
\end{equation}
where we used the fact that $p^i(\chi^{i}|Z^{i,-}_k) = p^i(\chi^{ij}_C|Z^{i,-}_k)\cdot p^i(\chi^{i\backslash j}|\chi^{ij}_C,Z^{i,-}_k)$. 
It can be seen that there are two contributions: the local pdf robot $i$ holds prior to fusion, and the marginal pdf it should receive from robot $j$ regarding their common variables $\chi_C^{ij}$, after removing the common data.  
This message can be viewed as sending a factor $f^{ji}(\chi_C^{ij})$ from robot $j$ to robot $i$, which is added into robot $i$'s factor graph as seen in Fig. \ref{fig:fusion}(d)-(e). 
The main questions that will be answered in the next section are \emph{how to design this message factor, and how to account for the common data in the denominator of (\ref{eq:Heterogeneous_fusion2})?}

The equations thus far hold for general distributions. However, to give more intuition, develop the theory, and be more explicit regarding the definition of factors and the conditional independence structure of the system, we will focus for the rest of this section on Gaussian distributions and linear operations.
Taking natural logarithm, (\ref{eq:Heterogeneous_fusion2}) can be written as,
\begin{equation}
    \begin{split}
        &\log p^i_f(\chi^i|Z^{i,+}_k) = \log p^i(\chi^{i}|Z^{i,-}_k)\\ 
        &+\log p^j(\chi^{ij}_C|Z^{j,-}_k) 
        -\log p_c^{ij}(\chi^{ij}_C|Z^{i,-}_k\cap Z^{j,-}_k)+\Tilde{C},
    \end{split}
    \label{eq:Log_Heterogeneous_fusion}
\end{equation}
where $\Tilde{C}$ is the normalization constant. 
Taking the first and second derivatives reveals the sufficient statistics in information form, namely the information vector ($\zeta$) and matrix ($\Lambda$), respectively. 
The message factor $f^{j\rightarrow i}(X_C^{ij})$ from robot $j$ to $i$, can then be defined in information form as,
\begin{equation}
    f^{j\rightarrow i}(\chi_C^{ij})=f^j(\chi^{ij}_C)-f_c^{ij}(\chi^{ij}_C),
    \label{eq:msgFactor}
\end{equation}
and 
\begin{equation}
    \begin{split}
        &f^j(\chi^{ij}_C) \propto p^j(\chi^{ij}_C|Z^{j,-}_k) \sim \mathcal{N}^{-1}(\bar{\zeta}^j_{\chi_C^{ij}}, \bar{\Lambda}^j_{\chi_C^{ij}\chi_C^{ij}})\\
        &f_c^{ij}(\chi^{ij}_C) \propto p_c^{ij}(\chi^{ij}_C|Z^{i,-}_k\cap Z^{j,-}_k) \sim \mathcal{N}^{-1}(\bar{\zeta}_{\chi_C^{ij}}^{ij}, \bar{\Lambda}^{ij}_{\chi_C^{ij}\chi_C^{ij}}).
    \end{split}
\end{equation}
Where $\bar{\zeta}_{\chi_C^{ij}}$ and $\bar{\Lambda}_{\chi_C^{ij}\chi_C^{ij}}$ are the marginal information vector and matrix respectively, and the superscripts $j$ and $ij$ refer to whether it taken from robot $j$'s pdf or the common pdf $p_c^{ij}(\cdot)$.
Note that while the marginal information vector and matrix, representing robot $j$'s pdf can be evaluated using
 \begin{equation}
     \begin{split}
         &\bar{\zeta}_{\chi_C^{ij}} = \zeta_{\chi_C^{ij}}-\Lambda_{\chi_C^{ij}\chi^{j\setminus i}}\Lambda^{-1}_{\chi^{j\setminus i}\chi^{j\setminus i}}\zeta_{\chi^{j\setminus i}},\\ &\bar{\Lambda}_{\chi_C^{ij}\chi_C^{ij}}=\Lambda_{\chi_C^{ij}\chi_C^{ij}}-\Lambda_{\chi_C^{ij}\chi^{j\setminus i}}\Lambda^{-1}_{\chi^{j\setminus i}\chi^{j\setminus i}}\Lambda_{\chi^{j\setminus i}\chi_C^{ij}},
         \label{eq:marg_statistics}
     \end{split}
 \end{equation}
the question of how to evaluate the common data factor $f_c^{ij}(\chi^{ij}_C)$ remains. 
In the next section, two approaches to evaluate $f_c^{ij}(\chi^{ij}_C)$ are detailed. 

\subsection{The Common Data Factor}
\label{subsec:commonFactor}
In the DDF literature, there are several methods to account for the common data, which can be divided into explicit and implicit methods.
In explicit methods, the transition of data through the system is either tracked by keeping a pedigree \cite{martin_distributed_2005} or, when the communication graph is undirected and acyclic, by adding a channel filter (CF) \cite{grime_data_1994} on the communication path between any two robots $i$ and $j$. 
The original CF explicitly calculates $p^{ij}_c(\chi|Z^{i,-}_k \cap Z^{j,-}_k)$ over the full (homogeneous) set of rvs $\chi$.
Implicit methods, e.g., covariance intersection (CI) \cite{julier_non-divergent_1997}, assume an unknown degree of dependency between the pdfs held by the robots, resulting from data common to the two robots. 
These methods then remove data from the pdf in a way that guarantees that data are only counted once but at the expense of ``throwing out" unique data. 

In the following, we explain how to compute the common data factor, $f_c^{ij}(\chi^{ij}_C)$, and the message factor, $f^{j\rightarrow i}(\chi^{ij}_C)$, for explicit and implicit heterogeneous fusion rules, which are heterogeneous extensions of the homogeneous CF and CI, as developed in \cite{dagan_exact_2023} and \cite{dagan_non-linear_2023}, respectively.



\subsubsection{\textbf{Heterogeneous Channel Filter (CF)}}
In \cite{dagan_exact_2023} we extend the idea of the homogeneous CF \cite{grime_data_1994} to heterogeneous DDF with the HS-CF. 
In HS-CF, each robot $i$ maintains a set of factor graphs $\mathcal{F}^{ij}_{CF}$, representing the CF between robot $i$ and every neighboring robot $j\in N^i_r$, over their common variables $\chi^{ij}_C$.
Similarly, the CF factor graph, $\mathcal{F}^{ji}_{CF}$, maintained by robot $j$ expresses its local estimate of the common marginal pdf $p^{ij}_c(\chi^{ij}_C|Z^{i,-}_k \cap Z^{j,-}_k)$ in the denominator of (\ref{eq:Heterogeneous_fusion2}). 
When robot $j$ forms a message to send to robot $i$, it computes the marginal over common variables from its local graph $\mathcal{F}^j$ and then removes the set of factors $f^{ij}_c(\chi^{ij}_C)=\prod_{l}f^{ij}_{c,l}(\chi^{ij}_{C,l})$ maintained by the CF factor graph $\mathcal{F}^{ji}_{CF}$.
Where the product is over all factors $l$ of subsets $\chi^{ij}_{C,l}$ of common variables between $i$ and $j$.
In problems where the pdfs are expressed using the marginal information vector ($\bar{\zeta}_{\chi_C^{ij}}$) and matrix ($\bar{\Lambda}_{\chi_C^{ij}\chi_C^{ij}}$), representing the mean and covariance of the pdf, the message factor over the subset of common rvs $\chi^{ij}_C$, in (\ref{eq:msgFactor}) is $f^{j\rightarrow i}(\chi^{ij}_C)\sim \mathcal{N}^{-1}(\chi^{ij}_C; \bar{\zeta}^{j\rightarrow i}_{\chi_C^{ij}}, \bar{\Lambda}^{j\rightarrow i}_{\chi_C^{ij}\chi_C^{ij}})$, with   
\begin{equation}
    \begin{split}
        &\bar{\zeta}^{j\rightarrow i}_{\chi_C^{ij}} = \bar{\zeta}^j_{\chi_C^{ij}}-\bar{\zeta}_{\chi_C^{ij},c}^{ij},\\ 
        &\bar{\Lambda}^{j\rightarrow i}_{\chi_C^{ij}\chi_C^{ij}} = \bar{\Lambda}^j_{\chi_C^{ij}\chi_C^{ij}}-\bar{\Lambda}^{ij}_{\chi_C^{ij}\chi_C^{ij},c}.
        \label{eq:marginalCF}
    \end{split}
\end{equation}

\subsubsection{\textbf{Heterogeneous Covariance Intersection (CI)}}
Covariance intersection \cite{julier_non-divergent_1997} computes the weighted average of the robots' information vector and matrix.
Ref. \cite{dagan_non-linear_2023} extends the homogeneous CI fusion rule to the heterogeneous HS-CI fusion rule by approximating the sufficient statistics of the `common' pdf $p^{ij}_c(\chi|Z^{i,-}_k \cap Z^{j,-}_k)$ with 
\begin{equation}
    \begin{split}
        &\bar{\zeta}_{\chi_C^{ij},c}^{ij} = (1-\omega)\bar{\zeta}^i_{\chi_C^{ij}}+\omega \bar{\zeta}^j_{\chi_C^{ij}},   \\
        &\bar{\Lambda}^{ij}_{\chi_C^{ij}\chi_C^{ij},c} = (1-\omega)\bar{\Lambda}^i_{\chi_C^{ij}\chi_C^{ij}}+\omega \bar{\Lambda}^j_{\chi_C^{ij}\chi_C^{ij}},
        \label{eq:marginalCI}
    \end{split}
\end{equation}
where the weight, $\omega$, is calculated to optimize some predetermined cost function, e.g., the trace or determinant of the fused covariance matrix. 

Substituting these definitions into (\ref{eq:marginalCF}), the information vector and matrix of CI message factor $f^{j\rightarrow i}(\chi^{ij}_C)$ are
\begin{equation}
    \begin{split}
        &\bar{\zeta}^{j\rightarrow i}_{\chi_C^{ij}} = (1-\omega) \bar{\zeta}^j_{\chi_C^{ij}}-
        (1-\omega)\bar{\zeta}^i_{\chi_C^{ij}},   \\ 
        &\bar{\Lambda}^{j\rightarrow i}_{\chi_C^{ij}\chi_C^{ij}} = (1-\omega )\bar{\Lambda}^j_{\chi_C^{ij}\chi_C^{ij}}-
        (1-\omega)\bar{\Lambda}^i_{\chi_C^{ij}\chi_C^{ij}}.
        \label{eq:msgFactorCI}
    \end{split}
\end{equation}
Note that prior to communication, robot $j$ doesn't hold an estimate of robot $i$'s information vector and matrix, and vice versa. 
In practice, the message factor is computed at the receiving robot end, i.e., after receiving $(\bar{\zeta}^j_{\chi_C^{ij}}, \bar{\Lambda}^j_{\chi_C^{ij}\chi_C^{ij}})$ from $j$, robot $i$ computes $\omega$ and adds the CI message factor (\ref{eq:msgFactorCI}) to its graph.

\section{The Conservative FG-DDF Algorithm}
\label{sec:algorithm}
In this section, we present the factor graph-based DDF algorithm \emph{FG-DDF}.
The pseudo-code of the algorithm is given in Algorithm \ref{algo:FG-DDF}, 
where we describe the steps robot $i$ takes for recursive filtering and heterogeneous peer-to-peer DDF with its $N_r^i$ neighbors by the form of graph operations.
For completeness, Appendix \ref{sec:appendix} details Kalman filtering-type operations on the graph, such as prediction, roll-up (marginalization), and measurement update.
We then detail the supporting fusion algorithms for sending and fusing a message in algorithms \ref{algo:sendMsg} and \ref{algo:fuseMsg}, respectively.
As described in Sec. \ref{subsec:commonFactor}, there are different methods to account for common data; while we detail the HS-CF and HS-CI, other heterogeneous fusion algorithms can be implemented instead (see \cite{dagan_exact_2023}). 
Since there are extra steps when using the CF algorithm, such as initializing a CF factor graph (Algorithm \ref{algo:FG-DDF}, line 5), we explicitly state the use of the CF option, otherwise it is assumed that the heterogeneous CI is used.

The rest of this section details our conservative filtering approach and algorithm (Algorithm \ref{algo:cons_filter}), and explains how inference is performed via a message-passing algorithm.


\begin{algorithm}[bt]
    \caption{FG-DDF}
    \label{algo:FG-DDF}
    \begin{algorithmic}[1]
    \State Define: $\chi_{i}$, Priors, Fusion algorithm (e.g., CF, CI)
    \State Initialize local factor graph $\mathcal{F}^{i}$
    \If{CF algorithm}\For{all $j \in N_r^i$}
    \State Initialize CF factor graph $\mathcal{F}^{ij}$ over $\chi_C^{ij}$
    \EndFor
    \EndIf 
    \For{all time steps}
    \For{all dynamic rvs $x_{k-1} \in \chi_{i}$}
    \State Add prediction nodes ($x_k$) and factors 
    \Comment{{\color{Gray} eq. \ref{eq:predFactors} }}
    \EndFor
    \State Conservative filtering $x_{k-1}$ \Comment{{\color{Gray} Algorithm \ref{algo:cons_filter} }}  
    \For{all measurements}
    \State Add measurement factors
    \Comment{{\color{Gray} eq. \ref{eq:measFactors} }}
    \EndFor
    \For{all $j \in N_r^i$}
    \State Send message to $j$ \Comment{{\color{Gray} Algorithm \ref{algo:sendMsg} }} 
    \State Fuse message from $j$
     \Comment{{\color{Gray} Algorithm \ref{algo:fuseMsg} }}   
    \EndFor
    \EndFor
    \State \Return
    \end{algorithmic}
\end{algorithm}

\begin{algorithm}[bt]
    \caption{Send Message to $j$}
    \label{algo:sendMsg}
    \begin{algorithmic}[1]
    \State {Input: $j$ - robot to communicate with} 
    \State Marginalize out non-mutual variables $\chi^{i\setminus j}$, receive marginal graph $\bar{\mathcal{F}}^i$ over $\chi^{ij}_C$    \Comment{{\color{Gray}  Algorithm \ref{algo:marginalize}}} 
    \If{CF algorithm}
    \State Subtract $\mathcal{F}_{CF}^{ij}$ from $\bar{\mathcal{F}}^i$ to get $f^{i\rightarrow j}(\chi^{ij}_C)$ \Comment{{\color{Gray}eq. \ref{eq:marginalCF}}}
    \State Update $\mathcal{F}_{CF}^{ij}$ with factors $f^{i\rightarrow j}(\chi^{ij}_C)$
    \Else
    \State Set $f^{i\rightarrow j}(\chi^{ij}_C)$ to factors of $\bar{\mathcal{F}}^i$
    \EndIf
    \State \Return Message $f^{i\rightarrow j}(\chi^{ij}_C)$
    \end{algorithmic}
\end{algorithm}

\begin{algorithm}[bt]
    \caption{Fuse Message from $j$ }
    \label{algo:fuseMsg}
    \begin{algorithmic}[1]
    \State {Input: In-message robot $j$ - ${f}^{j\rightarrow i}(\chi_C^{ij})$, Out-message robot $i$ - ${f}^{i\rightarrow j}(\chi_C^{ij})$} 
    \If{CF algorithm}
    \For{all $f \in {f}^{j\rightarrow i}(\chi_C^{ij})$}
    \State Add $f$ to local factor graph $\mathcal{F}^{i}$
    \State Add $f$ to CF factor graph $\mathcal{F}^{ij}_{CF}$
    \EndFor
    \Else
    \State Compute weight $\omega$ based on ${f}^{j\rightarrow i}(\chi_C^{ij})$ and ${f}^{i\rightarrow j}(\chi_C^{ij})$
    \State Compute implicit common data \Comment{{\color{Gray}eq. \ref{eq:marginalCI}}}
    \State Add to local factor graph $\mathcal{F}^{i}$
    \EndIf
    \State \Return
    \end{algorithmic}
\end{algorithm}

\begin{algorithm}[bt]
    \caption{Marginalize out variable $x$ from factor graph $\mathcal{F}$}
    \label{algo:marginalize}
    \begin{algorithmic}[1]
    \State {Input: variable to remove $x$, factor graph $\mathcal{F}$} 
    \State Sum $f(x)$ and all factors $f(x,\bar{x}_i)$ adjacent to $x$
    \State Create new marginal factor $f(\bar{x})$ using (\ref{eq:marginalFactor})
    \State Add edges from every $\bar{x}_i \in \bar{x}$ to $f(\bar{x})$
    \State Remove $f(x,\bar{x}_i)$ factor and $x$ from $\mathcal{F}$
    \State \Return marginal factor graph $\bar{\mathcal{F}}$
    \end{algorithmic}
\end{algorithm}


\subsection{Conservative Filtering}
\label{subsec:consFiltering}

The problem of conservative filtering is rooted in the dependencies caused by the marginalization of the time-dependent nodes \cite{dagan_conservative_2022}. 
This results in the loss of conditional independence between sets of locally relevant variables and breaks one of the basic assumptions that allow for heterogeneous fusion. 
As we show next, our solution includes two main steps, based on the based on the nature, or category, of the dependency: 
(i) A 'hidden' dependency is a dependency between subsets of non-mutual variables, thus some of the variables are hidden from the robot's point of view, i.e. they do not exist in its graph.
(ii) A 'visible' dependency is between subsets of common variables. In this case, all variables exist in the robot's graph and thus are visible to it.

Before detailing our analysis and solution for the heterogeneous filtering problem, it is important to define the meaning of conservative in the context of this paper.
Informally, Lubold and Taylor \cite{lubold_formal_2021} claim that a conservative fused posterior pdf should ``overestimate the uncertainty of a system”.  
The question regarding the more formal meaning in the case of general pdfs is beyond the scope of this paper and the reader is referred to \cite{lubold_formal_2021} and \cite{dagan_exact_2023} for further discussion.
Later in this paper, when a practical implementation is developed (Sec. \ref{subsec:deflation}) it will be assumed that the distributions can be described using their first two moments (mean and covariance). 
In this case, the commonly used definition of conservative states the difference between the fused covariance ($\Sigma$) and the covariance of a centralized estimator ($\Sigma^{cent}$) is positive semi-definite (PSD), that is $\Sigma-\Sigma^{cent}\succeq 0$. 
Here $\succeq$ means PSD, but in the more general case of comparing two pdfs we will use $\succeq$ to denote conservative.

\textit{Hidden dependencies:} Consider a two-robot problem, with non-mutual subsets of variables $\chi^{i\setminus j}=\chi_L^i$ and $\chi^{j\setminus i}=\chi_L^j$ and a common subset $\chi_C^{ij}$. 
Fig. \ref{fig:hiddenDep} shows an example of hidden dependency due to filtering, demonstrated on robot $i$'s local graph.
Fig. \ref{fig:hiddenDep}(a) shows robot $i$'s graph after the first relative measurement -- fusion -- prediction steps, and corresponds to the joint pdf,
\begin{equation}
    p(\chi^i_{2:1}|Z^{i,+}_{1}) = p(\chi_{C,1}^{ij},\chi^i_L|Z^{i,+}_{1})\cdotp(\chi_{C,2}^{ij}|\chi_{C,1}^{ij},Z^{i,+}_{1}).
    \label{eq:jointPrior}
\end{equation}

\input{Figures/hiddenDependencies}

Then marginalizing $\chi_{C,1}^{ij}$ (Fig. \ref{fig:hiddenDep}(b)) results in coupling shown by the red factor, which is \emph{hidden} from robot $i$'s perspective ($\chi_L^j$ does not exist in its local graph) but is evident in the full graph. 
To avoid this hidden coupling robot $i$ needs to take a preventative action based on its local understanding of the distribution. 
This is done by a two-step operation, as shown in Fig. \ref{fig:hiddenDep}(c)-(d): (i) decoupling the common variable $\chi_{C,1}^{ij}$ and the local variable $\chi_L^i$ by separating the factor $f_{y_1^i}$ to two unary factors (c);
(ii) marginalizing out $\chi_{C,1}^{ij}$.
Notice that now the local variables $\chi_L^i$ and $\chi_L^j$ are again conditionally independent given the common target $\chi_{C,2}^{ij}$ as shown in (d). 
These graph operations correspond to the following approximation and marginalization of $p(\chi^i_{2:1}|Z^{i,+}_{1})$ given in (\ref{eq:jointPrior}), respectively, 
\begin{equation}
    \begin{split}
        \tilde{p}(\chi^i_{2:1}|Z^{i,+}_{1}) = p(\chi_{C,1}^{ij},\chi_{C,2}^{ij})\cdot p(\chi^i_L),
    \end{split}
    \label{eq:local_k21}
\end{equation}
\begin{equation}
    \begin{split}
        \tilde{p}(\chi^i_{2}|Z^{i,+}_{1}) = p(\chi_{C,2}^{ij})\cdot p(\chi^i_L).
    \end{split}
    \label{eq:local_k1}
\end{equation}
Note that while the idea here is similar to what is done in graph-SLAM by the \emph{Exactly Sparse Extended Information Filter} (ESEIF) \cite{walter_exactly_2007} during sparsification, the difference is in the fact that the variables in our case are hidden and the correlations can't be set to zero directly. 

\textit{Visible dependencies:} 
Consider a chain-structured network with three robots, $j-i-m$. 
The common variables are again separated into three different sets $\chi_C^{ijm}$, $\chi_C^{ij\setminus m}$ and $\chi_C^{im\setminus j}$ -- variables common to the three robots, common to $i$ and $j$ but not to $m$ and similarly common to $i$ and $m$ but not to $j$, respectively. 

A priori, the sets $\chi_C^{ij\setminus m}$ and $\chi_C^{im\setminus j}$
are conditionally independent as seen in Fig. \ref{fig:common_vars}(a) and is given by the following pdf,
\begin{equation}
    \begin{split}
        &p(\chi^i_{2:1}|Z^{i,+}_{1})=\\
        &p(\chi_L^i,\chi_{C,2:1}^{ijm})\cdot 
        p(\chi_{C,2:1}^{ij\setminus m}|\chi_{C,1}^{ijm})\cdot p(\chi_{C,2:1}^{im\setminus j}|\chi_{C,1}^{ijm}).
    \end{split}
    \label{eq:initial_pdf}
\end{equation}
\input{Figures/Common_variables_fig}
The problem is that after the marginalization of past nodes (e.g., time step 1, Fig. \ref{fig:common_vars}(b)-(c)), the common sets of robot $i$ are all dependent, as shown by the red dense factor.
The key idea is to approximate the dense graph by a sparse graph such as to regain the original conditional independence structure (Fig. \ref{fig:common_vars}(d)), 
\begin{equation}
    \begin{split}
        &\hat{p}(\chi^i_{2}|Z^{i,+}_{1})=\\
        &p(\chi_L^i)\cdot p(\chi_{C,2}^{ijm})\cdot 
        p(\chi_{C,2}^{ij\setminus m}|\chi_{C,2}^{ijm})\cdot p(\chi_{C,2}^{im\setminus j}|\chi_{C,2}^{ijm}).
    \end{split}
    \label{eq:final_approximation}
\end{equation}
The last step is to increase the uncertainty of the approximated sparse pdf $\hat{p}(\cdot)$ such that,
\begin{equation}
    \begin{split}
        \hat{p}(\chi^i_{2}|Z^{i,+}_1)\succeq  p(\chi^i_{2}|Z^{i,+}_1),
    \end{split}
    \label{eq:pdfConservative}
\end{equation}
where, as noted before, $\succeq$ denotes conservative, and $p(\cdot)$ is the dense pdf that would have resulted without the approximations,
\begin{equation}
    \begin{split}
        p(\chi^i_{2}|Z^{i,+}_1)= \int p(\chi^i_{2:1}|Z^{i,+}_1)d\chi_{C,1}^i.
    \end{split}
    \label{eq:densepdf}
\end{equation}
While no assumption on the type of distribution has been made until now, there is no commonly used formal definition of conservativeness for general pdfs. 
Thus we now focus our attention on the case where the pdf can be represented by a Gaussian distribution, or their first two moments (mean and covariance). 
These are used in many applications across robotics \cite{dellaert_factor_2021}.

\begin{algorithm}[bt]
    \caption{Conservative Filtering}
    \label{algo:cons_filter}
    \begin{algorithmic}[1]
    \State \textbf{Input:} Local factor graph $\mathcal{F}^{i}$ describing $p(\chi^i_{k}|Z^{i,+}_{k-1})$
    \State Create a copy of the `true' graph $\mathcal{F}^{i}_{tr}$ 
    \State Approximate $\mathcal{F}^{i}$ with marginal pdfs \Comment{{\color{Gray} Eq. (\ref{eq:local_k21}) }}
    \State Marginalize out past nodes in $\mathcal{F}^{i}$ (\ref{eq:local_k1}) and $\mathcal{F}^{i}_{tr}$
    \State Regain conditional independence in $\mathcal{F}^{i}$ \Comment{{\color{Gray} Eq. (\ref{eq:final_approximation}) }}
    \State Guarantee conservativeness w.r.t  $\mathcal{F}^{i}_{tr}$ \Comment{{\color{Gray} Eq. (\ref{eq:sparse_cons_mat}) }}
    \For{Every neighbor $j\in N_r^i$ }
    \State Update CF graph $\mathcal{F}^{ij}_{CF}$
    \EndFor
    \State Add measurement nodes for time step $k$
    \State \Return
    \end{algorithmic}
\end{algorithm}

\subsection{Conservative Deflation}
\label{subsec:deflation}
Consider the case where both the dense (\ref{eq:densepdf}) and sparse (\ref{eq:pdfConservative}) pdfs are described by a factor graph with Gaussian factors, represented by an information vector and matrix.
We wish to make the approximate sparse Gaussian pdf, $\hat{p}(\chi^i_{2}|Z^{i,+}_1)\sim \mathcal{N}(\zeta_{sp},\Lambda_{sp})$, shown in Fig. \ref{fig:common_vars}(d) and given in (\ref{eq:final_approximation}), conservative relative to the dense pdf $p(\chi^i_{2}|Z^{i,+}_1)\mathcal{N}(\zeta_{de},\Lambda_{de})$ (\ref{eq:densepdf}) in the positive semi-definite (PSD) sense,
\begin{equation}
    \begin{split}
        \Lambda_{de}-\Lambda_{sp}\succeq 0.
    \end{split}
    \label{eq:consFilter}
\end{equation}
Due to its relative simplicity and the fact that it does not require optimization, we choose to use the method suggested by \cite{forsling_consistent_2019} and generalized in \cite{dagan_exact_2023}.
Briefly, we solve for the deflation constant $\lambda_{min}$, the minimal eigenvalue of $\Tilde{Q}=\Lambda_{sp}^{-\frac{1}{2}}\Lambda_{de}\Lambda_{sp}^{-\frac{1}{2}}$ and enforce the mean of the sparse pdf to equal the mean of the dense pdf $\mu_{de}$. The conservative approximate pdf is then,
\begin{equation}
    \begin{split}
        \hat{p}(\chi^i_{2}|Z^{i,+}_1)\sim \mathcal{N}^{-1}(\lambda_{min}\Lambda_{sp}\mu_{de}, \lambda_{min}\Lambda_{sp}).
    \end{split}
    \label{eq:sparse_cons_mat}
\end{equation}


\subsection{Inference}

In the previous sections, we described how a robot builds a local factor graph over its variables of interest through a series of predictions, local measurements, and the fusion of data from neighboring robots. 
At this point, it is important to describe how, given the local factor graph, the inference is performed. 
The goal of inference, in this case, is to deduce the marginal pdfs of each variable of interest from the joint function described by the factor graph. 
When the factor graph is acyclic, a message-passing algorithm, i.e. the \emph{sum-product algorithm} \cite{frey_factor_1997}, \cite{kschischang_factor_2001} can be used to directly work on the factor graph. 
However, when there is more than one variable in common between two communicating robots, the post-fusion graph has cycles (loops) due to the marginalization of the local variables at the ``sending" robot $j$.
While algorithms for cyclic graphs exist \cite{frey_revolution_1997}, they are not guaranteed to converge to the correct solution \cite{plarre_extended_2004}.

To solve this problem, we suggest transforming the loopy factor graph into an acyclic graph by forming cliques over the cycles.
Note that we are not transforming the graph into a clique graph or a junction tree, but only forming the minimum number of cliques that will result in an acyclic graph as we explain next. 
It is worth mentioning here that Kaess \textit{et al.} \cite{kaess_bayes_2011} uses the elimination algorithm to transfer a factor graph into a Bayes tree for incremental updates and inference. 
This approach might be useful in the future but is not advantageous in the scope of this work, as for dynamic problems it focuses on filtering solutions while maintaining the conditional independence structure (see earlier in this section).
Thus instead, we can the algorithm simpler and keep the graph as a factor graph and only form cliques from the variables that are in loops and then summarize their factors to new joint factors, connected to the clique. 
This is demonstrated in Fig. \ref{fig:clique_fg}, where assuming an acyclic network $j-i-m$, the common variables are separated as before into three different sets $\chi_C^{ijm}$, $\chi_C^{ij\setminus m}$ and $\chi_C^{im\setminus j}$.
We can see that fusion results in loops in the graph. 
We restore the acyclic structure by forming a clique over the separation sets $\chi_L^i \cup \chi_C^{ijm}$. 
Note that this needs to be done only before inference and not for fusion or filtering.  

\input{Figures/cliqueGraph}


%% file: Figures/newFgDDF.tex
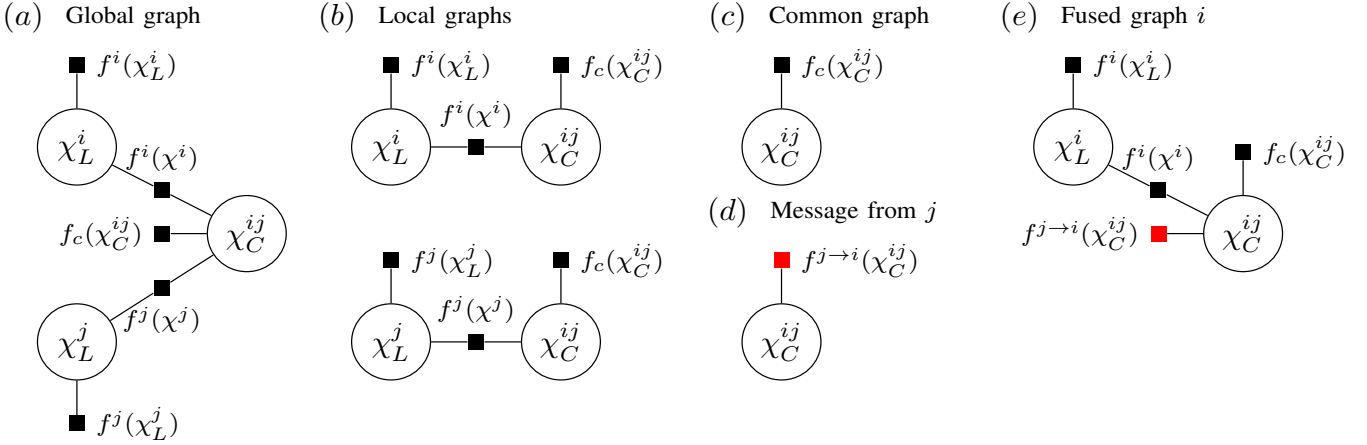
\begin{figure*}[bt]
\resizebox{7.2in}{2.4in}{%
\begin{tikzpicture}[ new set=import nodes]
 \begin{scope}[nodes={set=import nodes}]

     \node (a) at (-1.1,3) [label=right:Global graph] {$(a)$ };
      \node (x_Li)[latent, minimum size=25pt] at (-0.5,1.5) {$\chi^i_L$};
      \node (x_Lj)[latent, minimum size=25pt] at (-0.5,-0.75) {$\chi^j_L$};
      
      \node [latent, right=of x_Li, yshift=-1cm, minimum size=25pt] (x_ij) {$\chi^{ij}_{C}$};
      
      \node [factor, between=x_Li and x_ij,label=$f^{i}(\chi^i)$ ] (yi21) {};
      
      \node [factor, between=x_Lj and x_ij,label=below:$f^{j}(\chi^j)$] (f_j11) {};
      
      \node [factor, above=of x_Li,label=right:$f^{i}(\chi^i_L)$ ] (flia) {};
      \node [factor, below=of x_Lj,label=right:$f^{j}(\chi^j_L)$ ] (flja) {};
      \node [factor, left=of x_ij,label=left:$f_c(\chi^{ij}_C)$ ] (fcija) {};

      \node (b) at (2.4,3) [label=right:Local graphs] {$(b)$};
      
      \node (x_Lib)[latent, minimum size=25pt] at (3.0,1.5) {$\chi^i_L$};
      \node (x_Ljb)[latent, minimum size=25pt] at (3.0,-0.75) {$\chi^j_L$};
      
      \node [latent, right=of x_Lib,  minimum size=25pt] (x_ijbi) {$\chi^{ij}_{C}$};
      \node [latent, right=of x_Ljb,  minimum size=25pt] (x_ijbj) {$\chi^{ij}_{C}$};
      
      \node [factor, between=x_Lib and x_ijbi,label=$f^{i}(\chi^i)$ ] (yi21b) {};
      
      \node [factor, between=x_Ljb and x_ijbj,label=above:$f^{j}(\chi^j)$] (f_j11b) {};
      
      \node [factor, above=of x_Lib,label=right:$f^{i}(\chi^i_L)$ ] (flib) {};
      \node [factor, above=of x_Ljb,label=right:$f^{j}(\chi^j_L)$ ] (fljb) {};
      \node [factor, above=of x_ijbi,label=right:$f_c(\chi^{ij}_C)$ ] (fcijb) {};
      \node [factor, above=of x_ijbj,label=right:$f_c(\chi^{ij}_C)$ ] (fcijbj) {};
     
     \node (c) at (6.75,3) [label=right:Common graph]{$(c)$ };
      
      \node (x_ijc)[latent, minimum size=25pt] at (7.35,1.5) {$\chi^{ij}_{C}$};
            
      \node [factor, above=of x_ijc,label=right:$f_c(\chi^{ij}_C)$ ] (fcijc) {};
     
     \node (d) at (6.75,0.75) [label=right:Message from $j$] {$(d)$};
      
      \node (x_ijd)[latent, minimum size=25pt] at (7.35,-0.75) {$\chi^{ij}_{C}$};
            
      \node [factor,fill=red!100, above=of x_ijd,label=right:$f^{j\rightarrow i}(\chi^{ij}_C)$ ] (fcijd) {};
     \node (e) at (10,3) [label=right: Fused graph $i$] {$(e)$ };
      
      \node (x_Lie)[latent, minimum size=25pt] at (10.6,1.5) {$\chi^i_L$};

      \node [latent, right=of x_Lie, yshift=-1cm, minimum size=25pt] (x_ije) {$\chi^{ij}_{C}$};

       \node [factor, between=x_Lie and x_ije,label=$f^{i}(\chi^i)$ ] (yi21e) {};    
      
      \node [factor, above=of x_Lie,label=right:$f^{i}(\chi^i_L)$ ] (flie) {};

      \node [factor, above=of x_ije,label=right:$f_c(\chi^{ij}_C)$ ] (flce) {};

      \node [factor, fill=red!100,left=of x_ije,label=left:$f^{j\rightarrow i}(\chi^{ij}_C)$  ] (fljie) {};

  \end{scope}
  
 \graph {
    (import nodes);
   
    
    {x_ij,x_Li}--yi21,  
    
    {x_Lj}--f_j11,
    
    {x_ij}--f_j11,
    {x_Li}--flia, {x_Lj}--flja,
    fcija--x_ij,
   
    {x_Lib,x_ijbi}--yi21b,
    {x_Ljb,x_ijbj}--f_j11b,
    x_ijbi--fcijb,
    x_Lib--flib,
    x_Ljb--fljb,
    x_ijbj--fcijbj,

    x_ijc--fcijc,

    x_ijd--fcijd,

    x_ije--{yi21e,flce},
    x_ije--fljie,
    {yi21e, flie}--x_Lie,
    
    };
    
\end{tikzpicture}
 }
\caption{DDF in factor graphs: a) full (centralized) factor graph showing the local variable sets of robots $i$ and $j$ and the common variables set. b) showing the local factor graphs over each robot's variables of interest. c) is the common graph describing the factors common to both robots. d-e) demonstrates the fusion operation; the message sent from $j$ with new information over the common variables (d) is then integrated into $i$'s local graph with a simple factor addition (e).}
\label{fig:fusion}
\vspace{-0.2in}
      
\end{figure*}

%% file: Figures/hiddenDependencies.tex
\begin{figure}[tb]
\begin{tikzpicture}[ new set=import nodes]
 \begin{scope}[nodes={set=import nodes}]
      
      \node (a) at (0,1.5) {$(a)$};
      \node (s1)[latent, minimum size=25pt] at (0,0.75) {$\chi_L^i$};
      \node (s2)[latent, dashed, minimum size=25pt] at (0,-0.75) {$\chi_L^j$};
      \node [latent, right=of s1, yshift=-0.75cm] (x1) {$\chi_{C,1}^{ij}$};
      \node [factor, right=of x1,  xshift=0.25cm, label={$f(\chi_{C,2}^{ij}|\chi_{C,1}^{ij})$}] (f1) {};
      \node [latent, right=of f1, xshift=-0.25cm] (x2) {$\chi_{C,2}^{ij}$};
      \node [factor, between=x1 and s1 ,label=$f_{y^i_{1}}$] (yi1) {};
      \node [factor,fill=black!50, between=x1 and s2 ,label=below:$f_{y^j_{1}}$] (yj1) {};
      
      \node (b) at (5.5,1.5) {$(b)$};
      \node (s3)[latent, minimum size=25pt] at (5.5,0.75) {$\chi_L^i$};
      \node (s4)[latent, dashed, minimum size=25pt] at (5.5,-0.75) {$\chi_L^j$};
      \node [latent, right=of s3, yshift=-0.75cm] (x3) {$\chi_{C,2}^{ij}$};
      \node [factor,fill=red!100, between=x3 and s3, yshift=-0.375cm, label=\textcolor{red!100}{{$f_{m}$}}]  (f2) {};
     
      \node (c) at (0,-1.75) {$(c)$};
      \node (s1c)[latent, minimum size=25pt] at (0,-2.5) {$\chi_L^i$};
      \node (s2c)[latent, dashed, minimum size=25pt] at (0,-4.5) {$\chi_L^j$};
      \node [latent, right=of s1c, yshift=-0.75cm, xshift=0.25cm] (x1c) {$\chi_{C,1}^{ij}$};
      \node [factor, right=of x1c, xshift=0.225cm ] (f1c) {};
      \node [latent, right=of f1c, xshift=-0.5cm] (x2c) {$\chi_{C,2}^{ij}$};
      \node [factor, fill=red!100,below=of s1c, label=below:\textcolor{red!100}{$f_{y_1^i}(\chi_L^i)$}, yshift=0.2cm] (yi1siC) {};
      \node [factor, fill=red!100, above=of x1c, xshift=-0.1cm, label=right:\textcolor{red!100}{$f_{y_1^i}(\chi_{C,1}^{ij})$}, xshift=0.1cm] (yi1x1C) {};
      
      \node (d) at (5.5,-1.75) {$(d)$};
      \node (s1d)[latent, minimum size=25pt] at (5.5,-2.5) {$\chi_L^i$};
      \node (s2d)[latent, dashed, minimum size=25pt] at (5.5,-4.5) {$\chi_L^j$};
      \node [latent, right=of s1d, yshift=-0.75cm] (x2d) {$\chi_{C,2}^{ij}$};
      \node [factor,fill=red!100, between=x2d and s1d, yshift=-0.375cm, label=above:\textcolor{red!100}{$\tilde{f}_{m}$}]  (f2d) {};
      \node [factor, below=of s1d, label=below:$f_{y_1^i}(\chi_L^i)$, yshift=0.2cm] (yi1siD) {};
     
  \end{scope}
  
 \graph {
    (import nodes);
   
    x1--x2, 
    {x1,s1}--yi1, {s2}--[dashed]yj1, {x1}--yj1,
    {s3,x3}--f2, s4--[dashed]f2,
   
    x1c--x2c, 
    s1c--yi1siC,  x1c--yi1x1C,
    
    x2d--f2d, 
    s1d--yi1siD,
    };
    
\end{tikzpicture}
\caption{Example of hidden dependency due to filtering, demonstrated on robot $i$'s local graph and neighborhood variables, unimportant factors for the example are not shown. Dashed lines and gray factors are hidden from agent $i$. (a) Graph before marginalization step. (b) Naive marginalization creates hidden dependencies. (c)-(d) The proposed approach to avoid hidden dependencies. 
We use the notation $f(\chi_{C,2}^{ij}|\chi_{C,1}^{ij})$ to express and emphasize conditional dependency between variables \cite{frey_extending_2002}. 
}
      \label{fig:hiddenDep}
\end{figure}
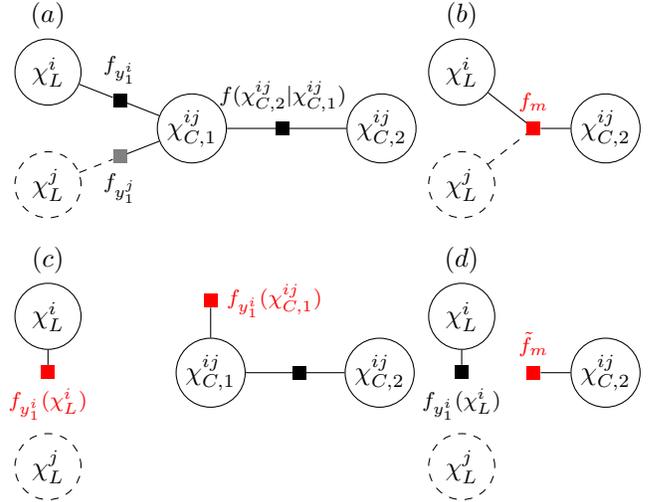

%% file: Figures/Common_variables_fig.tex
\begin{figure*}[bt!]
\resizebox{7.2in}{2.1in}{%
\begin{tikzpicture}[ new set=import nodes]
 \begin{scope}[nodes={set=import nodes}]

     \node (a) at (-0.35,3) {$(a)$};
      \node (x_Li)[latent, minimum size=31pt] at (-0.1,0) {$\chi^i_L$};
      \node (x_Lm)[latent,dashed, minimum size=31pt] at (0.6,2.25) {$\chi^m_L$};
      \node (x_Lj)[latent,dashed, minimum size=31pt] at (0.6,-2.25) {$\chi^j_L$};
      \node [latent, right=of x_Li, yshift=+1.5cm, minimum size=31pt] (x_im) {$\chi^{im\setminus j}_{C,1}$};
      \node [latent, right=of x_Li, yshift=0, minimum size=31pt] (x_ijm) {$\chi^{ijm}_{C,1}$};
      \node [latent, right=of x_Li, yshift=-1.5cm, minimum size=31pt] (x_ij) {$\chi^{ij\setminus m}_{C,1}$};
      \node [factor, between=x_Li and x_im,label=$f^{i,1}_{1}$ ] (yi11) {};
      \node [factor, between=x_Li and x_ijm,label=$f^{i,2}_{1}$ ] (yi21) {};
      \node [factor, between=x_Li and x_ij,label=$f^{i,3}_{1}$ ] (yi31) {};
     
      \node [factor, between=x_im and x_ijm ,label=right:$f^{im}_1$] (f_im) {};
      \node [factor, between=x_ij and x_ijm ,label=right:$f^{ij}_1$] (f_ij) {};
     
     \node [factor, right=of x_im,  label={$f^{im}_{2|1}$}] (f2_im) {};
      \node [latent, right=of f2_im, xshift=-0.65cm, minimum size=31pt] (x2_im) {$\chi^{im\setminus j}_{C,2}$};
      
      \node [factor, right=of x_ijm,   label={$f^{ijm}_{2|1}$}] (f2_ijm) {};
      \node [latent, right=of f2_ijm, xshift=-0.65cm, minimum size=31pt] (x2_ijm) {$\chi^{ijm}_{C,2}$};
      
      \node [factor, right=of x_ij, label={$f^{ij}_{2|1}$}] (f2_ij) {};
      \node [latent, right=of f2_ij, xshift=-0.65cm, minimum size=31pt] (x2_ij) {$\chi^{ij\setminus m}_{C,2}$};
      
      \node [factor,fill=black!40, between=x_Lj and x_ijm] (f_j11) {};
      \node [factor,fill=black!40, between=x_Lj and x_ij] (f_j21) {};
      
      \node [factor,fill=black!40, between=x_Lm and x_ijm] (f_m11) {};
      \node [factor,fill=black!40, between=x_Lm and x_im] (f_m21) {};

      \node (b) at (5.15,3) {$(b)$};
      \node (x_Lia)[latent, minimum size=31pt] at (5.5,0) {$\chi^i_L$};
      \node (x_Lma)[latent,dashed, minimum size=31pt] at (6.0,2.25) {$\chi^m_L$};
      \node (x_Lja)[latent,dashed, minimum size=31pt] at (6.0,-2.25) {$\chi^j_L$};

      \node [latent, right=of x_Lia, xshift=0.0cm, minimum size=31pt] (x2_ijma) {$\chi^{ijm}_{C,2}$};
      
      \node [latent, above=of x2_ijma, yshift=-0.5cm, minimum size=31pt] (x2_ima) {$\chi^{im\setminus j}_{C,2}$};
      
      \node [latent, below=of x2_ijma, yshift=0.5cm, minimum size=31pt] (x2_ija) {$\chi^{ij\setminus m}_{C,2}$};
     
     \node [factor,fill=red!100, right=of x2_ijma ] (f2a) {};
     \node [factor,fill=red!100, above=of x_Lia ] (fla) {};
     
      \node (c) at (9.25,3) {$(c)$};
      \node (x_Lib)[latent, minimum size=31pt] at (9.80,0) {$\chi^i_L$};
      \node (x_Lmb)[latent,dashed, minimum size=31pt] at (10.1,2.25) {$\chi^m_L$};
      \node (x_Ljb)[latent,dashed, minimum size=31pt] at (10.1,-2.25) {$\chi^j_L$};

      \node [latent, right=of x_Lib, xshift=0.0cm, minimum size=31pt] (x2_ijmb) {$\chi^{ijm}_{C,2}$};
      
      \node [latent, above=of x2_ijmb, yshift=-0.5cm, minimum size=31pt] (x2_imb) {$\chi^{im\setminus j}_{C,2}$};
      
      \node [latent, below=of x2_ijmb, yshift=0.5cm, minimum size=31pt] (x2_ijb) {$\chi^{ij\setminus m}_{C,2}$};
     
     \node [factor,fill=red!100, right=of x2_ijmb ] (f2b) {};
     \node [factor,fill=red!100, above=of x_Lib ] (flb) {};
     
     \node [factor,fill=black!40, between=x_Ljb and x2_ijmb] (f_j11c) {};
     \node [factor,fill=black!40, between=x_Ljb and x2_ijb] (f_j21c) {};
      
     \node [factor,fill=black!40, between=x_Lmb and x2_ijmb] (f_m11c) {};
     \node [factor,fill=black!40, between=x_Lmb and x2_imb] (f_m21c) {};
     
     \node [factor, between=x_Lib and x2_imb ] (yi11c) {};
     \node [factor, between=x_Lib and x2_ijmb ] (yi21c) {};
     \node [factor, between=x_Lib and x2_ijb ] (yi31c) {};
     
      
      \node (d) at (13.35,3) {$(d)$};
      \node (x_Lic)[latent, minimum size=31pt] at (13.9,0) {$\chi^i_L$};
      \node (x_Lmc)[latent,dashed, minimum size=31pt] at (14.2, 2.25) {$\chi^m_L$};
      \node (x_Ljc)[latent,dashed, minimum size=31pt] at (14.2,-2.25) {$\chi^j_L$};

      \node [latent, right=of x_Lic, xshift=0.0cm, minimum size=31pt] (x2_ijmc) {$\chi^{ijm}_{C,2}$};
      
      \node [latent, above=of x2_ijmc, yshift=-0.5cm, minimum size=31pt] (x2_imc) {$\chi^{im\setminus j}_{C,2}$};
      
      \node [latent, below=of x2_ijmc, yshift=0.5cm, minimum size=31pt] (x2_ijc) {$\chi^{ij\setminus m}_{C,2}$};
     
     \node [factor,fill=red!100, between=x2_ijmc and x2_imc, label=right:\textcolor{red!100}{{$\tilde{f}^{im}_{2}$}} ] (f2c_im) {};
     \node [factor,fill=red!100, between=x2_ijmc and x2_ijc, label=right:\textcolor{red!100}{{$\tilde{f}^{ij}_{2}$}} ] (f2c_ij) {};
     
     \node [factor,fill=red!100, above=of x_Lic ] (flc) {};
     
      \node (e) at (17.45,3) {$(e)$};
      \node (x_Lid)[latent, minimum size=31pt] at (18,0) {$\chi^i_L$};
      \node (x_Lmd)[latent,dashed, minimum size=31pt] at (18.3,2.25) {$\chi^m_L$};
      \node (x_Ljd)[latent,dashed, minimum size=31pt] at (18.3,-2.25) {$\chi^j_L$};

      \node [latent, right=of x_Lid, xshift=0.0cm, minimum size=31pt] (x2_ijmd) {$\chi^{ijm}_{C,2}$};
      
      \node [latent, above=of x2_ijmd, yshift=-0.5cm, minimum size=31pt] (x2_imd) {$\chi^{im\setminus j}_{C,2}$};
      
      \node [latent, below=of x2_ijmd, yshift=0.5cm, minimum size=31pt] (x2_ijd) {$\chi^{ij\setminus m}_{C,2}$};
     
     \node [factor,fill=red!100, between=x2_ijmd and x2_imd, label=right:\textcolor{red!100}{{$\tilde{f}^{im}_{2}$}} ] (f2d_im) {};
     \node [factor,fill=red!100, between=x2_ijmd and x2_ijd, label=right:\textcolor{red!100}{{$\tilde{f}^{ij}_{2}$}} ] (f2d_ij) {};
     
     \node [factor,fill=red!100, above=of x_Lid ] (fld) {};
     
     \node [factor,fill=black!40, between=x_Ljd and x2_ijmd] (f_j11d) {};
     \node [factor,fill=black!40, between=x_Ljd and x2_ijd] (f_j21d) {};
      
     \node [factor,fill=black!40, between=x_Lmd and x2_ijmd] (f_m11d) {};
     \node [factor,fill=black!40, between=x_Lmd and x2_imd] (f_m21d) {};
     
     \node [factor, between=x_Lid and x2_imd ] (yi11d) {};
     \node [factor, between=x_Lid and x2_ijmd ] (yi21d) {};
     \node [factor, between=x_Lid and x2_ijd ] (yi31d) {};
      
  \end{scope}
  
 \graph {
    (import nodes);
   
    
    {x_im,x_Li}--yi11, 
    {x_ijm,x_Li}--yi21,  
    {x_ij,x_Li}--yi31,
    {x_ij,x_ijm}--f_ij,
    {x_im,x_ijm}--f_im,
    {x2_ij,x_ij}--f2_ij,
    {x2_im,x_im}--f2_im,
    {x2_ijm,x_ijm}--f2_ijm,
    {x_Lm}--[dashed]f_m21, 
    {x_Lm}--[dashed]f_m11,
    {x_Lj}--[dashed]f_j21, 
    {x_Lj}--[dashed]f_j11,
    {x_im}--[dashed]f_m21,
    {x_ijm}--[dashed]f_m11,
    {x_ij}--[dashed]f_j21, 
    {x_ijm}--[dashed]f_j11,
   
    
    {x2_ima, x2_ija, x2_ijma }--f2a,
    {x_Lia }--fla,
    
    {x_Lib,x2_ijmb}--yi21c,
    {x_Lib,x2_imb}--yi11c,
    {x_Lib,x2_ijb}--yi31c,
    {x2_imb, x2_ijb, x2_ijmb }--f2b,
    {x_Lmb}--[dashed]f_m21c, 
    {x_Lmb}--[dashed]f_m11c,
    {x_Ljb}--[dashed]f_j21c, 
    {x_Ljb}--[dashed]f_j11c,
    {x2_imb}--[dashed]f_m21c,
    {x2_ijmb}--[dashed]f_m11c,
    {x2_ijb}--[dashed]f_j21c, 
    {x2_ijmb}--[dashed]f_j11c,
    {x_Lib }--flb,

    {x2_ijmc, x2_imc}--f2c_im,
    {x2_ijmc, x2_ijc}--f2c_ij,
    {x_Lic }--flc,
    
    {x_Lid,x2_ijmd}--yi21d,
    {x_Lid,x2_imd}--yi11d,
    {x_Lid,x2_ijd}--yi31d,
    {x_Lid }--fld,
  
    {x_Lmd}--[dashed]f_m21d, 
    {x_Lmd}--[dashed]f_m11d,
    {x_Ljd}--[dashed]f_j21d, 
    {x_Ljd}--[dashed]f_j11d,
    {x2_imd}--[dashed]f_m21d,
    {x2_ijmd}--[dashed]f_m11d,
    {x2_ijd}--[dashed]f_j21d, 
    {x2_ijmd}--[dashed]f_j11d,
    {x2_ijmd, x2_imd}--f2d_im,
    {x2_ijmd, x2_ijd}--f2d_ij,
    
    };
    
\end{tikzpicture}
 }
\caption{(a) Visible dependencies with full graph perspective. (b) graph after accounting for hidden dependencies and filtering; (c) addition of measurement factors and dependencies at time step 2, red dense factor breaks conditional independence assumption required for heterogeneous fusion. (d)-(e) our proposed method to regain conditional independence by factorizing into smaller local factors.}
      \label{fig:common_vars}
      
\end{figure*}
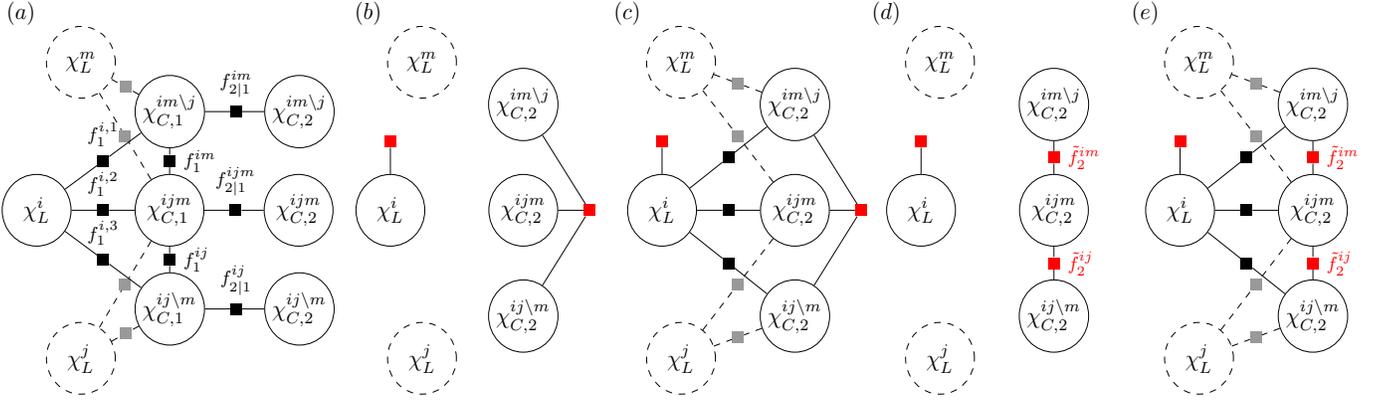

%% file: Figures/cliqueGraph.tex
\begin{figure}[bt]
\begin{tikzpicture}[ new set=import nodes]
 \begin{scope}[nodes={set=import nodes}]

     \node (a) at (-0.35,3) {$(a)$};
      \node (x_Li)[latent, minimum size=31pt] at (-0.1,0.5) {$\chi^i_L$};
      
      \node [latent, right=of x_Li, xshift=0.0cm, minimum size=31pt] (xk_ijm) {$\chi^{ijm}_{C,k}$};
      
      \node [latent, above=of xk_ijm, yshift=-0.5cm, minimum size=31pt] (xk_im) {$\chi^{im\setminus j}_{C,k}$};
      
      \node [latent, below=of xk_ijm, yshift=0.5cm, minimum size=31pt] (xk_ij) {$\chi^{ij\setminus m}_{C,k}$};     
     \node [factor, above=of x_Li ] (fl) {};
          
     \node [factor, between=x_Li and xk_im ] (yi11) {};
     \node [factor, between=x_Li and xk_ijm ] (yi21) {};
     \node [factor, between=x_Li and xk_ij ] (yi31) {};
      
  \end{scope}
  
 \graph {
    (import nodes);
   
     {x_Li,xk_ijm}--yi21,
    {x_Li,xk_im}--yi11,
    {x_Li,xk_ij}--yi31,
    {x_Li }--fl,
    
    };

    \begin{scope}[nodes={set=import nodes}]

     \node (b) at (3.5,3) {$(b)$};
      \node (x_Li)[latent, minimum size=31pt] at (3.75,0.5) {$\chi^i_L$};
      
      \node [latent, right=of x_Li, xshift=0.0cm, minimum size=31pt] (xk_ijm) {$\chi^{ijm}_{C,k}$};
      
      \node [latent, above=of xk_ijm, yshift=-0.5cm, minimum size=31pt] (xk_im) {$\chi^{im\setminus j}_{C,k}$};
      
      \node [latent, below=of xk_ijm, yshift=0.5cm, minimum size=31pt] (xk_ij) {$\chi^{ij\setminus m}_{C,k}$};     
     \node [factor, above=of x_Li ] (fl) {};
          
     \node [factor, between=x_Li and xk_im ] (yi11) {};
     \node [factor, between=x_Li and xk_ijm ] (yi21) {};
     \node [factor, between=x_Li and xk_ij ] (yi31) {};

     \node [factor,fill=red!100, between=xk_ijm and xk_im, label=right:\textcolor{red!100}{{${f}^{im}_{k}$}} ] (f2_im) {};
     \node [factor,fill=red!100, between=xk_ijm and xk_ij, label=right:\textcolor{red!100}{{${f}^{ij}_{k}$}} ] (f2_ij) {};
      
  \end{scope}
  
 \graph {
    (import nodes);
   
     {x_Li,xk_ijm}--yi21,
    {x_Li,xk_im}--yi11,
    {x_Li,xk_ij}--yi31,
    {x_Li }--fl,
    {xk_ijm,xk_im}--f2_im,
    {xk_ijm,xk_ij}--f2_ij,

    };

    \begin{scope}[nodes={set=import nodes}]

     \node (c) at (-0.35,-2.25) {$(c)$};
      \node (xk_im)[latent, minimum size=31pt] at (1,-2.75) {$\chi^{im\setminus j}_{C,k}$};
      
      \node [ right=of xk_im, xshift=0.0cm] (xk_ijm) {$\chi_L^i\bigcup\chi^{ijm}_{C,k}$};
      
      \draw[red] (3.5, -2.75) ellipse [x radius=1, y radius=0.5 ];
      
      \node [latent, right=of xk_ijm, minimum size=31pt] (xk_ij) {$\chi^{ij\setminus m}_{C,k}$};     
          
     \node [factor, fill=red!100, between=xk_im and xk_ijm, xshift=-0.25cm ] (yi11) {};
     \node [factor, fill=red!100, between=xk_ijm and xk_ij, xshift=0.25cm ] (yi21) {};
     \node at (2.65,-2.75) (e1) {};
     \node at (4.35,-2.75) (e2) {};

      
  \end{scope}
  
 \graph {
    (import nodes);
   
     {xk_ij}--yi21,
    {xk_im}--yi11,
    e1--yi11,e2--yi21,

    };
    
\end{tikzpicture}
\caption{Transitioning the graph into a clique factor graph demonstrated on a network $j-i-m$. a) Local graph before fusion - acyclic. b) Messages sent from $k$ and $j$ to $i$ result in loops in the local graph. c) Acyclic structure is regained by forming a clique over separation sets.  } 
      \label{fig:clique_fg}
      \vspace{-0.2in}
      
\end{figure}
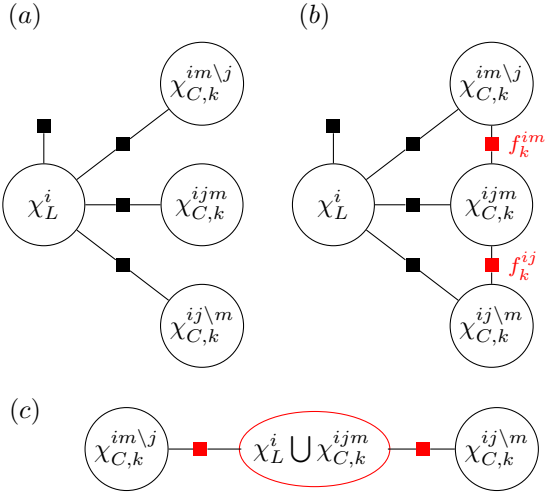

%% file: Text/6_Experiments.tex
To test and evaluate the proposed methods we performed both simulations and hardware experiments.
With Monte Carlo (MC) simulations of a multi-agent multi-target tracking application, we show that our algorithm results in a consistent and conservative estimate when compared to a centralized estimator. 
We then show the applicability of the algorithm with a large-scale nonlinear cooperative localization (CL) simulation in a cyclic network topology. 
Lastly, we show the robustness of the approach to message dropouts, hardware implementation, and non-matching dynamic models (i.e. when the model and reality do not match).
In all experiments, we use the FG-DDF Algorithm \ref{algo:FG-DDF}, with the addition of the conservative filtering Algorithm \ref{algo:cons_filter}. 
Each robot then maintains and reasons over its local dynamic factor graph and calculates its local MMSE estimate (mean and covariance) using the sum-product (message passing) algorithm on its factor graph \cite{frey_factor_1997}.

\subsection{Simulations}
\input{Text/6p1_Simulation}

\label{subsec:Sim}

\subsection{Hardware Experiments}
\input{Text/6p2_hardwareExp}

\label{subsec:Exp}

%% file: Text/6p1_Simulation.tex
\subsubsection*{\textbf{Consistency and Conservativeness}}
\label{subsec:simA}
We demonstrate the consistency and conservativeness of the proposed methods a Monte Carlo tracking simulation involving 4 robots tracking 6 dynamic targets.
The robots are connected in an acyclic topology $(1\leftrightarrow 2\leftrightarrow 3\leftrightarrow 4)$ with bidirectional communication,
which enables the use of the heterogeneous CF (HS-CF).
Each robot is tasked with estimating the $2D$ position and velocity $x^t=[n^t,\dot{n}^t,e^t,\dot{e}^t]^T$ of a subset of the 6 targets, and its own constant (but unknown) robot-to-target relative position measurement bias $s^i=[b^i_{n},b^i_{e}]^T$, similar to the bias in \cite{noack_treatment_2015}. 
At time step $k$, any robot $i\in N_r$ can take two types of measurements, a relative measurement to target $t$, $y^{i,t}_{k}$,  and a measurement to a known landmark, $m^i_{k}$,
\begin{equation}
    \begin{split}
        &y^{i,t}_{k} = x^t_k+s^i+v^{i,1}_k, \ \ v^{i,1}_k \sim \mathcal{N}(0,R^{i,1}),  \\
        &m^i_{k} = s^i+v^{i,2}_k, \ \ v^{i,2}_k \sim \mathcal{N}(0,R^{i,2}).
    \end{split}
    \label{eq:meas_model}
\end{equation}
In all simulations and experiments, unless otherwise stated, a measurement is taken at each time step. 

The robots' target tracking and self-localization (bias estimation) tasks are described by their local random state vector:
\begin{equation}
    \begin{split}
        &\chi^1_k=\begin{bmatrix} x^1_k \\ x^2_k \\ x^3_k \\ s^1 \end{bmatrix},
        \chi^2_k=\begin{bmatrix} x^2_k \\ x^3_k \\ s^2 \end{bmatrix},
        \chi^3_k=\begin{bmatrix} x^2_k \\ x^3_k \\ x^4_k \\ x^5_k  \\ s^3 \end{bmatrix},
        \chi^4_k=\begin{bmatrix} x^4_k \\ x^5_k \\ x^6_k \\ s^4 \end{bmatrix},\\
        &R^{1,1}=R^{1,2}=diag([1,5]), \ R^{2,1}=R^{2,2}=diag([3,3]), \\
        &R^{3,1}=R^{3,2}=diag([4,4]), \ R^{4,1}=R^{4,1}=diag([5,1]).
    \end{split}
    \label{eq:tasks}
\end{equation}
For example, robot 1 is tasked with tracking targets 1--3 and its own local bias. 
Common states between robots are those states in the intersection between the state vectors, e.g., $\chi^{12}_{C,k} = [(x^2_k)^T, (x^3_k)^T]^T$ are the common random state vectors between robots 1 and 2, and $\chi^1_{L,k}=[(x^1_k)^T, (s^1)^T]^T$ are the local random state vectors of robot 1. 
Notice that in homogeneous fusion, all 4 robots must reason and communicate the full global set of 32 states, while in heterogeneous fusion robots only reason over their local tasks, which goes up to a maximum of 18 states (robot 3), and communicate over a maximum of 8 common states. 
For Gaussian distributions, this translates to more than $90\%$ reduction in communication and computation costs for robot 1,2,4, and about $82\%$ for robot 3.

Target $t$'s dynamics are modeled using a nearly constant velocity kinematic model, commonly used for tracking problem 
\cite{bar-shalom_linear_2001}, 
\begin{equation}
    \begin{split}
    &x^t_{k+1}=Fx^t_{k}+Gu^t_k+\omega_k, \ \ \omega_k \sim \mathcal{N}(0,0.08\cdot I_{n_x\times n_x}),\\
    &F=\begin{bmatrix}1 & \Delta t &0 &0\\0 &1 &0 &0\\ 0 &0 &1 & \Delta t\\0& 0 &0 &1 \end{bmatrix}, \quad
    G=\begin{bmatrix}\frac{1}{2}\Delta t^2 &0\\\Delta t&0\\0 &\frac{1}{2}\Delta t^2\\0 &\Delta t \end{bmatrix}.
    \end{split}
    \label{eq:dynamicEq}
\end{equation}
Note that this model is also used in the hardware experiments, where the true target dynamics are highly non-linear.

\begin{figure}[bt]
    \centering
    \includegraphics[width=0.48\textwidth]{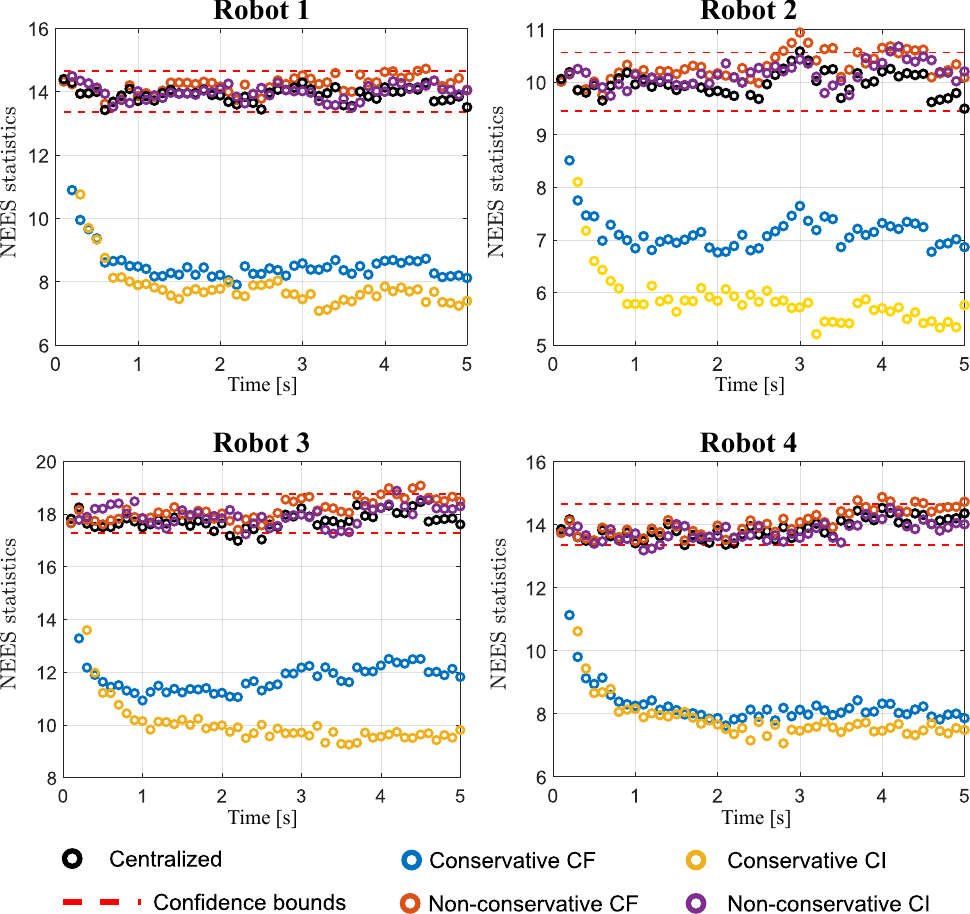}
    \caption{Multi-robot multi-target tracking NEES results. Showing comparison between the different fusion methods and a centralized estimator.}
    \label{fig:NEES_4a6t}
\end{figure}

We evaluate the FG-DDF performance by comparing the HS-CF and HS-CI algorithms, with and without the conservative filtering algorithm described in Sec. \ref{subsec:consFiltering}, to a centralized estimator's estimate over the marginal estimate of relevant states.
The consistency of each robot's estimate was tested using the normalized estimation error squared (NEES) test \cite{bar-shalom_linear_2001} over 250 MC simulations.  
Fig. \ref{fig:NEES_4a6t} shows the four robots' results with $95\%$ confidence bounds for all the robots. 
Results show that all robots are consistent, whereas, for the cases with conservative filtering, the robots underestimate their uncertainty due to the information matrix deflation (covariance inflation).

\begin{figure}[bt]
    \centering
    \includegraphics[width=0.48\textwidth]{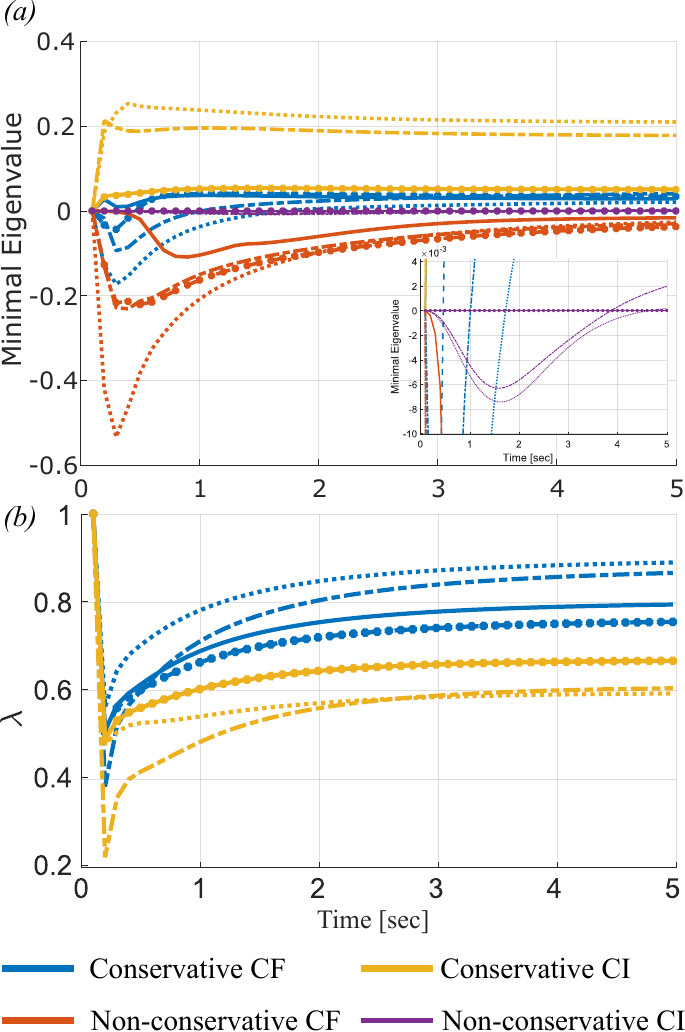}
    \caption{(a) Minimal eigenvalue of the covariance difference, positive values imply conservative result. The inset shows a zoomed view of the non-conservative CI. (b) Deflation constant comparison between CF and CI. Different line styles indicate different robots.  }
    \label{fig:minEval_4a6t}
\end{figure}

To validate that the robots' estimates are conservative, the local uncertainty of each robot was compared with the estimate of a centralized estimator. 
We require the robot's covariance to be larger than the centralized covariance in the PSD sense, $\Sigma_{\chi^i}-\Sigma_{\chi^i}^{cent} \succeq 0$, where $\Sigma_{\chi^i}^{cent}$ is the marginal covariance over the local random state vector of robot $i$, $\chi^i$, taken from the joint centralized covariance over the full system state $\chi$. 
In practice, this is verified by computing the minimal eigenvalue of the above covariance difference and testing that it is not negative. 
Fig. \ref{fig:minEval_4a6t}(a) shows the minimal eigenvalue of the covariance difference with and without the conservative filtering algorithm. 
For the HS-CF algorithm, it can be seen that without conservative filtering (orange) results are not conservative.
On the other hand, with the algorithm (blue), it takes between $0.5-2$ seconds for the robots' estimates to become conservative, but once it is conservative, it stays conservative. 
For the HS-CI algorithm, the estimates are conservative with the conservative filtering algorithm (yellow) and are just below the zero line without the algorithm (purple).
This non-conservative result demonstrates the importance of the conservative filtering approach and the difference from homogeneous fusion, as the homogeneous CI is guaranteed to provide conservative fusion results only if all robots estimate all global states in the problem \cite{julier_non-divergent_1997}. 

Fig. \ref{fig:minEval_4a6t}(b) shows the change in the deflation constant $\lambda$ in time across all robots for both the HS-CF (blue) and HS-CI algorithms (yellow) with conservative filtering. 
Intuitively, as robots accumulate more data, approximations by detaching dependencies have a lower impact on the pdf, i.e., the sparse approximation is `closer' to the dense pdf.
Thus $\lambda$ approaches some limit, depending on the problem statistics and structure. 
The three main insights are: 
(i) with the CF, across all robots, $\lambda$ is larger, meaning that more information is kept relative to the CI $\lambda$.
(ii) The effect of the measurement statistics on robots' $\lambda$ is different between the two methods. 
For CF, lower noise (robots 2,3) results in a larger $\lambda$, while for CI it is the opposite. 
During fusion in CI the information matrices are averaged which causes another deflation of the matrices. This results in weaker dependencies between variables during conservative filtering.
Since robots 2 and 3 have two fusion events, versus one for robots 1 and 4, 
they now accumulate less data and need a smaller constant to regain conservativeness.  
(iii) Both the minimal eigenvalues and $\lambda$ are constant across simulations. 
We attribute it to the fact that these are functions of the problem statistics, i.e., measurement and dynamic model noise, and of the communication network, all where kept constant across simulations. 
Thus in general, these can be analyzed and studied a priori, and are not dependent on the actual measurements received.  
However, we do expect them to change in nonlinear simulations for example, where the information matrix is dependent on the current estimate due to linearization.

\subsubsection*{\textbf{Applicability and Robustness}}
To demonstrate the applicability of our approach to real-world problems, we consider the problem of multi-robot nonlinear cooperative localization.
Over the past 30 years, there has been a plethora of work in CL suggesting different approaches and algorithms.
These mainly differ in their definition of (i) which states are estimated by each robot and (ii) how they account for the dependencies between robots' pose estimates, resulting from the relative robot-to-robot measurements, where they either augment the state or ``decorrelate" agents in an ad-hoc manner.
These methods exist along a spectrum of how robots account for state-dependency, then on the one end, there are algorithms such as the EKF-based CL \cite{roumeliotis_distributed_2002} where each robot estimates the full state of a team of $n_r$ robots and accounts for their induced dependencies. 
On the other end, there is the CI-CL algorithm \cite{carrillo-arce_decentralized_2013}, where each robot only maintains an estimate of its own (ego) pose and accounts for its dependency on other robots' poses, resulting from the relative measurement using CI. 
Our FG-DDF approach can be placed in different places across this spectrum -- when the robots have `dense' neighborhoods and take relative measurements to a large number of robots in the network, it will approach the first end (closer to \cite{roumeliotis_distributed_2002}), but when they have a `sparse' neighborhoods, with a limited number of neighbors, it will approach the second end of the spectrum (closer to \cite{carrillo-arce_decentralized_2013}).
Since the FG-DDF is designed to improve scalability, we will test it in a more sparse environment and compare it to the CI-CL algorithm \cite{carrillo-arce_decentralized_2013}, due to its simplicity of implementation and proven conservativeness.
It is important to note that our goal here is not to benchmark a new decentralized CL algorithm but to demonstrate the applicability of the FG-DDF to another real-world robotic application. 

\begin{figure}[tb]
    \centering
    \includegraphics[]{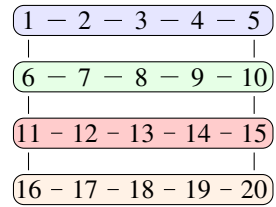}
    \caption{Undirected and cyclic network topology, split into 4 groups of 5 robots. Neighboring robots can communicate and take relative measurements of each other. }
    \label{fig:net_topology}
\end{figure}

We now turn our attention to the simulation scenario. 
In the CL scenario presented here, we simulate a cyclic network of 20 robots connected according to Fig. \ref{fig:net_topology}. 
In this cyclic setting, the conditional independence assumption is no longer $100\%$ valid, but if the cycles are big enough, the dependencies between far-away robots should be close to conditionally independent, and as we show below the FG-DDF algorithm provides good performance.

There are 3 main parts occurring in every time step: dynamics, measurements, and fusion.\\
\textbf{Dynamics:} The robots follow a nonlinear Dubin`s car model, 
\begin{equation}
    \begin{split}
        &\dot{x}^i=v^i\cos\theta^i+\omega^i_x,\\
        &\dot{y}^i=v^i\sin\theta^i+\omega^i_y,\\
        &\dot{\theta}^i = \frac{v}{L}\tan\phi^i+\omega^i_\theta,
    \end{split}
\end{equation}
where $x^i$, $y$, and $\theta^i$ are the robot $i$'s $2D$ positions and heading angle, respectively. $v^i$ and $\phi^i$, and $\omega^i=[\omega^i_x, \omega^i_y, \omega^i_\theta]^T$ are the time-dependent linear velocity ($m/s$), steering angle (rad), and zero mean additive white Gaussian noise (AWGN), respectively. 
$L$ is the front-rear wheel distance (taken to be $1$ m in the simulations).\\
\textbf{Measurements:}
At every time step, robots take local sensor bearing and range measurements with respect to 3--4 known landmarks and relative measurements to their neighbors, according to the following model,
\begin{equation}
    \begin{split}
        &h_{\theta}=tan^{-1}\left(\frac{y^t-y^i}{x^t-x^i} \right)-\theta^i+v^i_{\theta,k}  \ v^i_{\theta,k} \sim \mathcal{N}(0,{\sigma_{\theta}}^2), \\
        &h_{r} = \sqrt{(x^t-x^i)^2+(y^t-y^i)^2}+v^i_{r,k},  \ v^i_{r,k} \sim \mathcal{N}(0,{\sigma^{i}_r}^2), 
    \end{split}
\end{equation}
where $(x^t,y^t)$ are the measured target or landmark $2D$ position, $\sigma_{\theta}=1$ deg, and $\sigma^{i}_r$ is randomly sampled for each robot from values of $2/4/6$ m with equal probability prior to the MC simulations.\\
\textbf{Fusion:} 
Depending on the chosen decentralized fusion algorithm, each robot $i$ either communicates its marginal estimate over common states to its neighbor $j$ and then performs heterogeneous fusion, or sends only its measurement-based estimate of robot $j$ and fuses it using the CI-CL algorithm \cite{carrillo-arce_decentralized_2013}.

\begin{figure}[bt]
    \centering
    \includegraphics[width=0.48\textwidth]{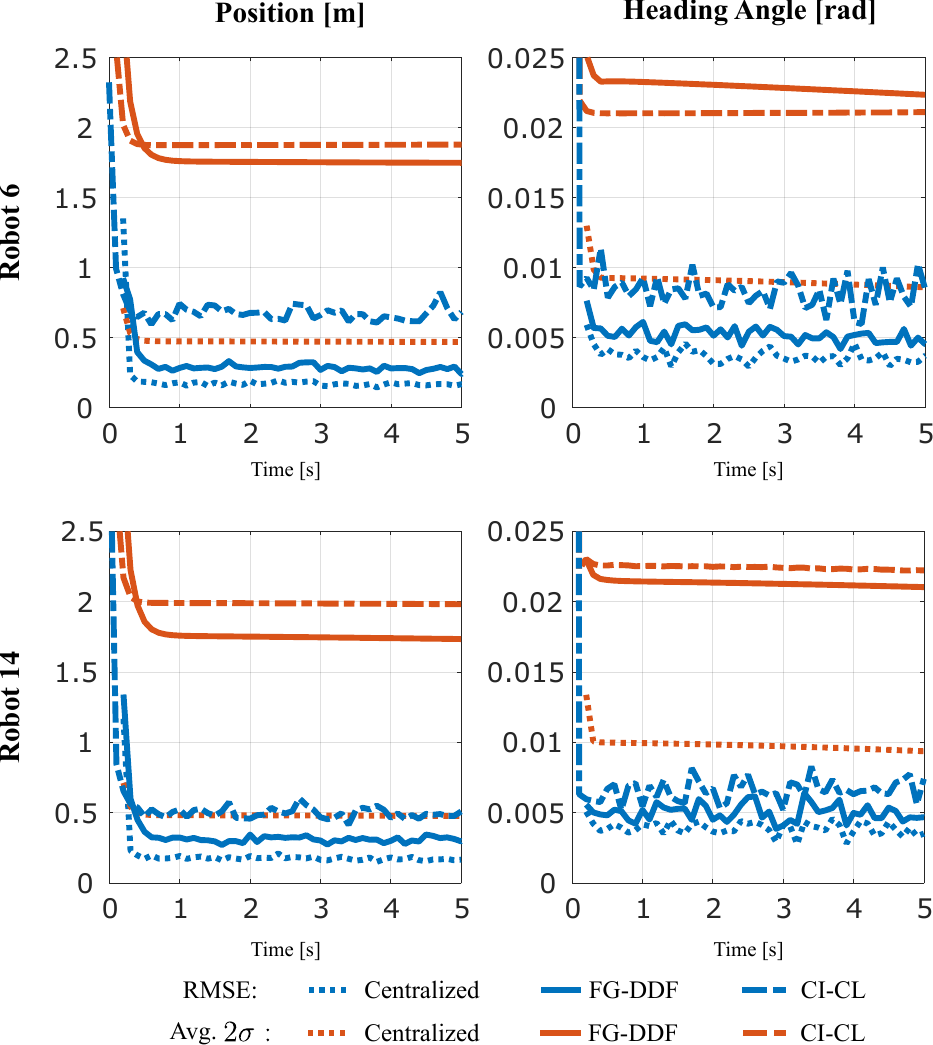}
    \caption{Representative results of CL MC simulation comparing centralized, FG-DDF, and CI-CL. Presented are robots 6 and 14 RMSE and average $2\sigma$ position and heading angle.}
    \label{fig:CL_results}
\end{figure}

The performance of the FG-DDF is compared to the CI-CL algorithm and to a centralized (global) estimator.
Table \ref{tab:CL_Results} compares the RMSE position and $1\sigma$ results, averaged across simulation time and over 50 MC runs. 
Shown are the results of the robots with the smallest (`best') and largest (`worst') RMSE out of the 20 robots, and the average across all robots. 
Fig. \ref{fig:CL_results} shows the 2D position and heading angle RMSE and $2\sigma$ results averaged across 50 MC simulations of the two representative robots with the smallest average position RMSE using FG-DDF (robot 6) and using CI-CL (robot 14).
As we see from the table and figure, FG-DDF results in a smaller RMSE and uncertainty than CI-CL over the robot's ego states, and, as expected, a larger RMSE and uncertainty compared to the centralized estimator.
To check the conservativeness of each robot ego estimate we again calculate the minimal eigenvalue of the difference between the marginal of a centralized estimator and the robot's local estimate ($\Sigma_{\chi^i}-\Sigma_{\chi^i}^{cent} \succeq 0$). 
For both approaches the minimal eigenvalues are close to zero, with positive (conservative) values for robots $1,4,7,10,13,16,19$, and negative eigenvalues (non-conservative) with values larger than $-0.06$ for all other robots.
Analyzing the eigenvectors corresponding to the minimal eigenvalues reveals that the overconfident direction corresponds to the heading angle.

\begin{table}[tb]
\renewcommand{\arraystretch}{1.2}
\caption{RMSE position error and $1\sigma$ [m].  }
    \begin{center}
    \begin{tabular}{c|c|c|c}
       Robot    & Centralized & FG-DDF - Ours  & CI-CL \cite{carrillo-arce_decentralized_2013}   \\ \hline
        Best & $0.19\pm0.24$ & \color{Green}{$0.31\pm0.89$} & $0.51\pm1.00$    \\ \hline
        Worst & $0.26\pm0.24$ & \color{Green}{$0.48\pm0.78 $} & $0.77\pm1.08$      \\ \hline
        Avg. & $0.22\pm0.24$&\color{Green}{$0.39\pm0.90$}& $0.63\pm0.99$  \\ \hline
    \end{tabular}
    \end{center}
    \label{tab:CL_Results}
    \vspace{-0.2in}
\end{table} 

We can attribute the FG-DDF's better estimation results to three key differences in the data or the estimate communicated between robots in the FG-DDF algorithm when compared to those communicated in the CI-CL algorithm:

\begin{enumerate}
    \item In FG-DDF, robot $i$ recursively updates its estimate of robot $j$'s pose, accounting for previous estimates (prior), thus it has `memory' of the estimate. 
    In CI-CL robot $i$ communicates an `ad-hoc' estimate, based only on its sensor characteristics and current estimate of its ego pose, thus it is `memoryless'.
    \item By augmenting robot $i$'s ego pose with neighboring robots, FG-DDF directly accounts for dependencies between the robots' states, while CI-CL purposely ignores these dependencies. 
    \item In FG-DDF, each relative robot-to-robot measurement, adds data to robot $i$'s ego estimate, while in CI-CL this data is only added to robot $j$'s ego estimate.
\end{enumerate}
It is important to note that the improved accuracy comes at the expense of 
an increase in communication and computation requirements when compared to the CI-CL approach. 
But, the increase computation is only proportional to the number of robot $i$'s neighbors $n^i_r$ and not the overall number of robots $n_r>n^i_r$, while the increase in communication is proportional to the number of common states robot $i$ has with $j$ ($|\chi^{ij}_C|\leq |\chi^i|$). 
For example, assume robot $i$ has two neighbors $j$ and $k$ to whom it takes relative measurements, and assume robots $j$ and $k$ are not neighbors.
Now, robot $i$ computation cost is $\mathcal{O}(3)$ and its communication cost is $\mathcal{O}(2)$.
This is compared to CI-CL's $\mathcal{O}(1)$ communication and computation costs.
The FG-DDF framework also allows for a `hybrid' approach, where robot $i$ can augment its ego pose with its `constant' neighbors (i.e. robots that are consistently in its neighborhood, e.g., $j$, $k$), and use CI-CL for 'random' neighbors that are occasionally in its neighborhood, further demonstrating the FG-DDF's flexibility in system design.

%% file: Text/6p2_hardwareExp.tex

\begin{figure}[htb!]
    \centering
    \includegraphics[width=0.48\textwidth]{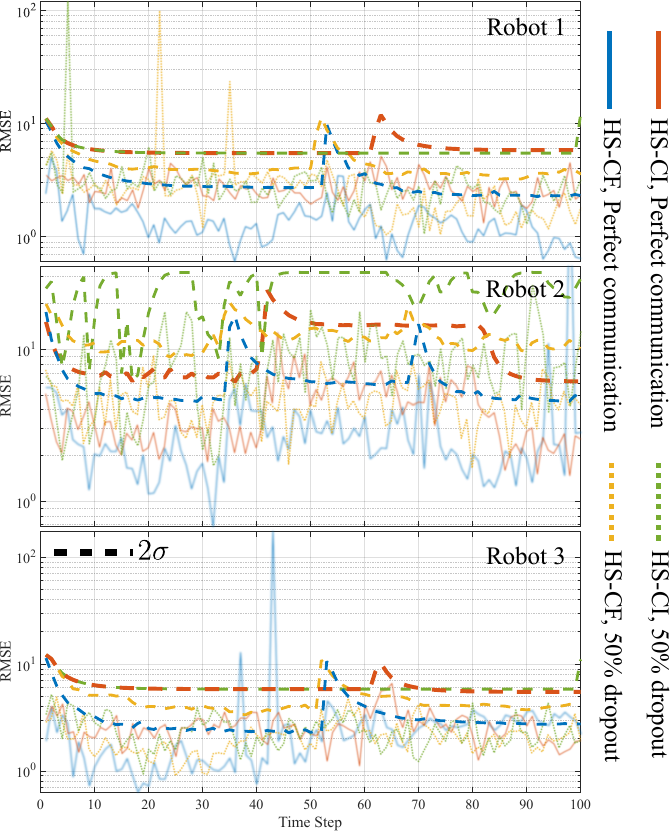}
    \caption{RMSE and $2\sigma$ of over each robot state vector \ref{eq:expTasks} from hardware experiments. Experiments tested HS-CF and HS-CI fusion algorithms under perfect communication and $50\%$ message dropout.}
    \label{fig:experimentResults}
    \vspace{-0.2in}
\end{figure}

To evaluate the robustness of the FG-DDF framework we:
(i) Implement our algorithms on hardware using three Clearpath Jackal UGVs as the trackers and 5 Adeept wheeled robots for Arduino (AWR-A) as targets (see Fig. \ref{fig:experimentResults}); 
(ii) Test under realistic conditions such as message dropout, measurement outliers, and model misalignment. 
The experiments include a target tracking and localization scenario, similar to the one presented in Sec. \ref{subsec:simA}, where the inference tasks of the robots are,
\begin{equation}
    \begin{split}
        \chi^1_k=\begin{bmatrix} x^1_k \\ x^2_k \\ s^1 \end{bmatrix},
        \chi^2_k=\begin{bmatrix} x^2_k \\ x^3_k \\ x^4_k \\ s^2 \end{bmatrix},
        \chi^3_k=\begin{bmatrix} x^4_k \\ x^5_k \\ s^3 \end{bmatrix}.
    \end{split}
    \label{eq:expTasks}
\end{equation}

The Jackals robots are equipped with a 2-core Intel Celeron G1840 CPU with 4GB of RAM and 128GB of disk drive storage and a 2-core Intel i7-7500U CPU with 32GB of RAM and 512GB of disk drive storage, respectively.
Each robot runs the FG-DDF onboard as the inference and fusion engines, where ROS (version 1) is used for message passing between the robots. 

\begin{figure}[tb]
    \vspace{0.05in}
    \centering
    \includegraphics[width=0.48\textwidth]{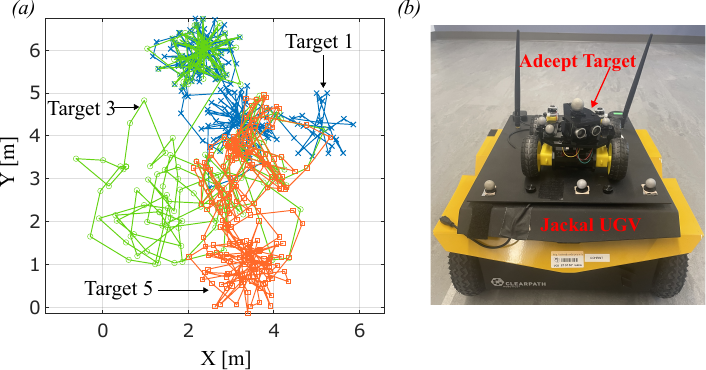}
    \caption{(a) 3 of the 5 targets' true trajectories as captured by the Vicon system. (b) picture of a tracker robot – Clearpath Jackal UGV, and a target – Adeept (AWR-A)}
    \label{fig:Targets_motion}
\end{figure}

In our experiments, as in many target tracking problems, the targets' dynamics are modeled using a linear `nearly constant velocity' motion model presented earlier in (\ref{eq:dynamicEq}). 
The linear relative target and landmark position measurements are gathered using Vicon motion-capture cameras, corrupted by zero mean Gaussian noise, and are modeled using (\ref{eq:meas_model}) with covariance values of $R^{i,1}=diag([1,5])$ and $R^{i,2}=diag([3,3])$.
In practice, The targets are programmed to move in a straight line for about 2 seconds and then turn right for half a second, but due to slipping, their turn angle varies stochastically, which results in a  nonlinear trajectory, resembling a random walk, as can be seen from the true trajectory in Fig. \ref{fig:Targets_motion}.


We performed 4 experiments, with the HS-CI and HS-CF fusion rules, each with a different communication success probability, and compared RMSE and $2\sigma$ bounds across each robot's states, based on truth values from the Vicon system\footnote{We assume that the data association problem is solved by a different algorithm.}. 
Figure \ref{fig:experimentResults} compares the RMSE and $2\sigma$ over each robot's local task (as defined in (\ref{eq:expTasks})) using the HS-CF and HS-CI algorithms for perfect and imperfect ($50\%$ dropout) communication, i.e. each message packet reaches its destination according to a Bernoulli distribution with $p=0.5$.

All robots perform well, where the error spike is attributed to outliers from the Vicon measurements, which occur when targets pass too close to each other. 
The $2\sigma$ lines show similar behavior to the one observed in simulations, where: 
(i) the HS-CF yields a more confident estimate than the HS-CI, i.e. tighter bounds; 
(ii) for the HS-CF the $50\%$ communication rate yields a worse estimate then perfect communication; 
(iii) for the HS-CI, message dropouts have indistinguishable effects on the estimates in this scenario for robots 1 and 3, while for robot 2 there is a significant difference between perfect communication (orange) and message dropout (green). 
We attribute this difference to the fact that robot 2 communicates with both robot 1 and 3, thus $50\%$ dropout means that only $25\%$ of the messages reach robot 2. 
These experiments demonstrate the robustness of the FG-DDF to real-world effects, such as message dropout and measurement outliers, and achieve good tracking performance, despite nonlinear target behavior.

%% file: Text/7_Conclusions.tex


In this paper, we developed a new architecture for multi-robot heterogeneous fusion.
Our solution frames the Bayesian decentralized data fusion (DDF) in terms of factor graphs, which has two main advantages. 
First, Bayesian DDF allows robots to share data without dependency on the robots' underlying sensors.
However, classical DDF does not scale, as it requires robots to share their full homogeneous state representation, which results in dependency on the size of the system, and not on the robot's task. 
Thus, the second advantage is that by using factor graphs, we can analyze and exploit the probabilistic conditional independence structure, inherent to muti-robot fusion problems, to: (i) split the global joint pdf, into smaller locally-relevant sub-graphs, and (ii) fuse data on only common subset of states. 
In practical settings such as cooperative tracking and localization with linear-Gaussian dynamics and measurement models, this can translate to more than $90\%$ communications and computation savings.

Results of simulations and hardware experiments validate the FG-DDF framework in real-world scenarios. 
These include non-linear dynamics and measurement models, imperfect communication, large cyclic networks, and measurement outliers.  
Our evaluation demonstrates the applicability and robustness of FG-DDF to different robotic applications.

%% file: Text/Appendix.tex
To complete the description of the FG-DDF approach, we provide a summary of the probabilistic operations that need to be performed on the graph during a Kalman filtering--type inference. 
These include three types of operations on the probabilistic graphical model: 
\emph{prediction}, \emph{roll-up} (marginalization of past states), and \emph{estimation} (measurement update). 
We follow \cite{paskin_thin_2002} and the information augmented state (\emph{iAS}) smoother presented in \cite{dagan_exact_2023} to define these operations on a factor graph and show their translation into new factors.
Since for Gaussian distributions, the factor graph is an equivalent representation of the information form, the factors below will be defined by an information vector and matrix $\{\zeta, \Lambda\}$

In the following we use $f(x_k|x_{k-1})=f(x_k,x_{k-1})\propto p(x_k|x_{k-1})$ and $f(x_k;y_k)=f(x_k)\propto l(x_k;y_k)$ to express and emphasize conditional probabilities between variables and between variables to measurements, respectively. 
This is based on the idea in \cite{frey_extending_2002} of extending factor graphs to represent directed (Bayesian networks) and undirected (Markov random fields), thus making dependencies explicit when reading the factor graph.

\textbf{Prediction:}
In the prediction step three factors are added to the graph: two unary factors $f^-(x_k), f^-(x_{k+1})$, connected to the variable nodes $x_k$ and $x_{k+1}$, respectively, and a binary factor $f^-(x_{k+1}|x_k)$ connected to both variables and describes the correlation between the two variables. 
The $(-)$ superscript denotes prediction.
\begin{equation}
    \begin{split}
        f^-(x_k) &= \{-F_k^TQ_k^{-1}G_ku_k,\ F_k^TQ_k^{-1}F_k\} \\
        f^-(x_{k+1}) &= \{Q_k^{-1}G_ku_k, \ Q_k^{-1} \} \\ 
        f^-(x_{k+1}|x_k) &= \left \{ \begin{pmatrix}
        0_{n\times1} \\ 0_{n\times1}
        \end{pmatrix}, \begin{pmatrix}
        0_{n\times n} & -Q_k^{-1}F_k\\ 
        -F_k^TQ_k^{-1} & 0_{n\times n}
        \end{pmatrix} \right \}.
    \end{split}
    \label{eq:predFactors}
\end{equation}
Here $F_k$ and $G_k$ are the state transition and control matrices at time step $k$, respectively. $u_k$ is the input vector and $Q_k$ is a zero mean white Gaussian process noise covariance matrix. The graphical description of the prediction step is given in Fig. \ref{fig:factorGraph_operations}b.

\input{Figures/fg_operations}

\textbf{Roll-up (marginalization):}
It is known that upon marginalization of a random variable $x$ in a directed graphical model, all variables in the Markov blanket of $x$ are moralized, i.e. ``married'' by adding an edge. 
The effect in a factor graph is similar, and variables are moralized by adding a factor connecting all variables in the Markov blanket of the marginalized variable. 
Denote the Markov blanket of $x$ by $\bar{x}$, then the new factor $f(\bar{x})$ is computed in two steps:\\
1. Sum all factors connected to $x$ to compute the following information vector and matrix:
\begin{equation}
    \{\zeta,\ \Lambda \} = f(x)+\sum_{i\in \bar{x}}f(x,\bar{x}_i)
\end{equation}
2. Use Schur complement to compute the marginal and find the new factor $f(\bar{x})$:
\begin{equation}
    f(\bar{x}) = \{\zeta_{\bar{x}}-\Lambda_{\bar{x}x}\Lambda_{xx}^{-1}\zeta_x,\  \Lambda_{\bar{x}\bar{x}}-\Lambda_{\bar{x}x}\Lambda_{xx}^{-1}\Lambda_{x\bar{x}}\}.
    \label{eq:marginalFactor}
\end{equation}
Notice that as a result, conditionally independent variables become correlated.\\
We demonstrate marginalization in Fig. \ref{fig:factorGraph_operations}c, marginalizing out $x_{k:n}$ induces a new factor $f(x_{k+1})$ over $x_{k+1}$. Here, since the only variable in the Markov blanket of the marginalized variables is $\bar{x}=x_{k+1}$, the new factor is unary over $x_{k+1}$ alone.

\textbf{Estimation (measurement update):}
Adding a measurement in the information form of the Kalman filter is a simple task as it includes only the variables of the current time step. 
In a factor graph this translates to adding a factor $f(x_{k+1};y_{k+1})$ connected to all measured variables,
\begin{equation}
    f(x_{k+1};y_{k+1}) = \{H_{k+1}^TR_{k+1}^{-1}y_{k+1},\ H_{k+1}^TR_{k+1}^{-1}H_{k+1} \}.
    \label{eq:measFactors}
\end{equation}
Where $H_{k+1}$ is the sensing matrix, $R_{k+1}$ is a zero mean white Gaussian measurement noise covariance matrix and $y_{k+1}$ is the noisy measurement vector.
Figure \ref{fig:factorGraph_operations}d shows the addition of a unary measurement factor $f(x_{k+1};y_{k+1})$.

%% file: Figures/fg_operations.tex
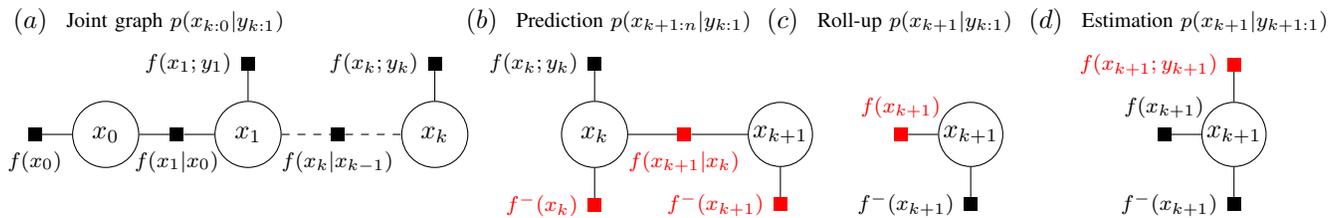
\begin{figure*}[tb]
\begin{tikzpicture}[ new set=import nodes]
 \begin{scope}[nodes={set=import nodes}]
      
      \node (a) at (-1.0,1.5) [label=right:Joint graph $p(x_{k:0}|y_{k:1})$] {$(a)$ };
      \node (x0)[latent, minimum size=25pt] at (0,0) {$x_0$};
      \node [latent, right=of x0, minimum size=25pt] (x1) {$x_1$};
      \node [factor, right=of x1,  xshift=0.25cm, label=below:{$f(x_k|x_{k-1})$}] (f1) {};
      \node [factor, left=of x0, label=below:$f(x_0)$]  (f0) {};
      \node [factor, above=of x1, label=left:$f(x_1;y_1)$]  (fy1) {};
      \node [latent, minimum size=25pt, right=of f1, xshift=-0.25cm] (xk) {$x_k$};
      \node [factor, between=x1 and x0 ,label=below:$f(x_1|x_0)$] (f01) {};
      \node [factor, above=of xk, label=left:$f(x_k;y_k)$]  (fyk) {};
      
      
      \node (c) at (9.0,1.5) [label=right:Roll-up $p(x_{k+1}|y_{k:1})$]{$(c)$};
      \node (xkp1c)[latent] at (11.5,0) {$x_{k+1}$};
      \node [factor, fill=red!100,left=of xkp1c,label=above:\textcolor{red!100}{$f(x_{k+1})$}] (fkp1c) {};
      \node [factor, below=of xkp1c, label=left:$f^-(x_{k+1})$]  (fkp1mc) {};
      
      \node (b) at (5.,1.5) [label=right:Prediction $p(x_{k+1:n}|y_{k:1})$]{$(b)$};

      \node (xkb)[latent, minimum size=25pt] at (6.5,0) {$x_k$};
      \node [factor, fill=red!100,  right=of xkb, xshift=0.25cm, label=below:\textcolor{red!100}{$f(x_{k+1}|x_{k})$}] (fk_kp1) {};
      \node [latent, right=of fk_kp1, xshift=-0.25cm] (xkp1b) {$x_{k+1}$};
            
      \node [factor, above=of xkb, label=left:$f(x_k;y_k)$]  (fykb) {};
      \node [factor,  fill=red!100,below=of xkb, label=left:\textcolor{red!100}{$f^-(x_k)$}]  (fkmb) {};
      \node [factor,  fill=red!100,below=of xkp1b, label=left:\textcolor{red!100}{$f^-(x_{k+1})$}]  (fkp1mb) {};
           
       \node (d) at (12.5,1.5) [label=right:Estimation $p(x_{k+1}|y_{k+1:1})$]{$(d)$};

      \node (xkp1d)[latent] at (15.0,0) {$x_{k+1}$};
      \node [factor, left=of xkp1d,label=above:{$f(x_{k+1})$}] (fkp1d) {};
      \node [factor, below=of xkp1d, label=left:$f^-(x_{k+1})$]  (fkp1md) {};
      \node [factor, fill=red!100, above=of xkp1d, label=left:\textcolor{red!100}{$f(x_{k+1};y_{k+1})$}]  (fykp1d) {};

  \end{scope}
  
 \graph {
    (import nodes);
    x1--[dashed]xk, x0--f0, 
    {x1,x0}--f01, x1--fy1, xk--fyk, 

    fkp1c--xkp1c, fkp1mc--xkp1c,

    xkp1b--fkp1mb, xkb--fykb, xkb--fkmb,
    {xkb,xkp1b}--fk_kp1,

    xkp1d--{fkp1d,fkp1md,fykp1d}
    };
    
\end{tikzpicture}
    \caption{Factor graph operations. (a) The full factor graph, including time history, running a sum-product algorithm is equivalent to a smoothing solution. (b)-(d) Represent a filtering approach with prediction, roll-up, and estimation steps, respectively. }
    \label{fig:factorGraph_operations}

\end{figure*}